\pdfoutput=1
\documentclass[10pt, a4paper]{article}

\usepackage[final]{lrec2026}

\usepackage[T1]{fontenc}
\usepackage[utf8]{inputenc}
\usepackage{microtype}
\usepackage{inconsolata}
\usepackage{graphicx}
\usepackage[dvipsnames]{xcolor}
\usepackage{CJKutf8}
\usepackage{subcaption}
\usepackage{tabularx}
\usepackage{booktabs}
\usepackage{caption}
\usepackage{subcaption}
\usepackage{listings}
\usepackage{hyperref}
\usepackage{arydshln}
\usepackage[inline]{enumitem}

\usepackage{tcolorbox}
\tcbuselibrary{listings, breakable}
\newtcblisting{promptbox}{
    listing only,
    breakable, % allows breaking across pages
    colback=gray!5,
    colframe=gray!40,
    boxrule=0.5pt,
    arc=6pt, % 0pt for square corners
    left=3pt,
    right=3pt,
    top=0pt,
    bottom=0pt,
    listing options={
        basicstyle=\small\ttfamily,
        breaklines=true,
        breakindent=0pt,
        breakatwhitespace=false,
        columns=flexible,
        keepspaces=true,
        showstringspaces=false,
        escapechar=@
    }
}

\usepackage{todonotes}
\newcommand{\yh}[1]{{\sethlcolor{yellow!35}\hl{#1}}}
\newcommand{\rh}[1]{{\sethlcolor{red!20}\hl{#1}}}

\newenvironment{myquote}%
  {\list{}{\leftmargin=0.1in\rightmargin=0.1in}\item[]}%
  {\endlist}

\title{A Longitudinal, Multinational, and Multilingual Corpus of News Coverage of the Russo-Ukrainian War}

\name{%
Dikshya Mohanty\textsuperscript{1},
Taisiia Sabadyn\textsuperscript{1},
Jelwin Rodrigues\textsuperscript{1},\\
\large\bf%
Chenlu Wang\textsuperscript{1},
Abhishek Kalugade\textsuperscript{1},
Ritwik Banerjee\textsuperscript{1,2}%
}

\address{%
\textsuperscript{1}Department of Computer Science, \textsuperscript{2}AI Innovation Institute\\
Stony Brook University, New York\\
\{dimohanty, jelrodrigues, chenlwang, rbanerjee\}@cs.stonybrook.edu\\
\{taisiia.sabadyn, abhishek.kalugade\}@stonybrook.edu
}

\abstract{%
We present \textsc{dnipro}, a corpus of 246K news articles from the Russo-Ukrainian war (Feb 2022 -- Aug 2024) spanning eleven outlets across five nation-states (Russia, Ukraine, U.S., U.K., China) and three languages. The corpus features comprehensive metadata and human-evaluated annotations for stance, sentiment, and topical framing, enabling systematic analysis of \textit{competing geopolitical narratives}.
It is uniquely suited for empirical studies of narrative divergence, media framing, and information warfare.
Our exploratory analyses reveal how media outlets construct incompatible realities through divergent attribution and topical selection without direct refutation of opposing narratives.
\textsc{dnipro} empowers empirical research on narrative evolution, cross-lingual information flow, and computational detection of implicit contradictions in fragmented information ecosystems.
\\
\newline
\mbox{}
}

\begin{document}
\maketitleabstract

\section{Introduction}
\label{sec:introduction}
The Russo-Ukrainian war, which escalated on February 24, 2022 with the declaration of a ``special military operation'' by the Russian government,\footnote{Within 48 hours, media in Russia were banned from describing the developments as ``war''~\cite{russia2024ban}.} has significantly impacted the global geopolitical landscape~\cite{dervis2023transformation, silvertsen2024projecting}.
The war has redrawn traditional alliances, tested the resolve of international institutions, and created new geopolitical fault lines.
Despite initial expectations of a swift Russian victory, Ukraine's resistance has resulted in a protracted war with substantial loss of life and human capital~\cite{watling2023preliminary, onemillion}.
This conflict has also been characterized by a quieter battle for narrative control, with news coverage from several nations playing a pivotal role in shaping global public opinion and influencing the war's trajectory~\cite{bradshaw2024strategic}.
These sources reflect their national priorities, strategic perspectives, and the perceptions they intend to create among information consumers~\cite{johais2024unleash}.
Yet, despite the centrality of media and narrative framing, there remains a lack of comprehensive datasets for longitudinal analysis of how nations have framed the war for their domestic and international audiences.

We introduce a collection of \textsc{\small \textbf{D}iverse \textbf{N}arratives and \textbf{I}nternational \textbf{P}erspectives on the \textbf{R}usso-Ukrainian \textbf{O}ffensive} (\textsc{dnipro}), a comprehensive longitudinal corpus of news coverage from February 1, 2022 to August 31, 2024 --- the longest time span of any multinational or multilingual news corpus on the Russia-Ukraine war. The code-base and supporting materials are publicly available.
The corpus encompasses eleven influential media outlets and news agencies aligned with five primary geopolitical actors with strategic interests in the conflict.
Following methodological approaches established by \citet{nygren2016journalism} and \citet{roman2017information}, the sources include media from direct participants---Russia and Ukraine---and influential global powers with economic, diplomatic, or military involvement: the USA, the UK, and China.
The inclusion of state-affiliated media from Russia and Ukraine captures opposing conflict narratives that play a pivotal role in shaping assessments under uncertain information~\cite{saletta2020}. 
Media from China, the USA, and the UK represent perspectives of nations with substantial geopolitical and economic interests in the conflict's trajectory, as well as the military, diplomatic, and economic leverage to influence its outcome~\cite{zhang2016china, thussu2017china}. 
These diverse perspectives and sources enable more comprehensive analysis of complex developments by helping analysts overcome the limitations of single-narrative framings~\cite{werd2018thesis}.

\textsc{dnipro} offers rich annotations for topical framing, entity-level stance, and entity-level sentiment, enabling event tracking and other \textit{longitudinal} analyses across geopolitical perspectives.
The dataset does not attempt to verify the accuracy of any claims; rather, it provides a resource to study how wartime narratives are constructed and disseminated by major geopolitical actors.\footnote{This aligns with established methodologies in discourse analysis~\cite{entman1993framing, fairclough2010critical} examining how language and framing construct social reality. \citet{gamson1989media} demonstrated how this approach reveals the competition shaping public understanding of contentious events.}
The corpus is uniquely suited to such studies, as it comprises 246{,}229 articles across 11 media sources and 3 languages, spanning 5 diverse nations, over a period of 31 months. Our \textbf{key contributions} include:
\begin{itemize}[leftmargin=17pt, nolistsep, noitemsep]
\item[(1)] a longitudinal, multinational, and multilingual corpus of news coverage spanning the first 31 months of the Russia-Ukraine war;
\item[(2)] inclusion of diverse media, enabling cross-cultural examination of conflict narratives; and
\item[(3)] annotations for comparative analyses of named entities, sentiment, stance, and topical framing.
\end{itemize}
Next, we place \textsc{dnipro} within the landscape of existing conflict-related corpora to highlight its distinct contributions to the field (\S\hspace{1pt}\ref{sec:related-work}), and describe how our design and construction choices (\S\hspace{1pt}\ref{sec:data-collection}) make it a powerful resource for large-scale analyses of wartime narratives.
We employ \textsc{dnipro} in several applications, using state-of-the-art natural language understanding systems and additional annotations (\S\hspace{1pt}\ref{sec:data-annotations} and \S\hspace{1pt}\ref{sec:applications}), demonstrating its utility.
%We conclude with a discussion of the limitations of contemporary approaches, and propose future directions to understand divergent narratives emerging from disparate sociocultural and national spheres.

% \textsc{dnipro} addresses significant gaps in systematic and longitudinal cross-platform coverage of the war, which exist despite the creation of several datasets: from large-scale Twitter archives and Reddit collections to region-specific news media studies. Existing social media datasets excel at capturing real-time public sentiment and polarization~\cite{fung2022weibo, haq2022twitter, shevtsov2022twitter, zhu2022reddit}. Albeit offering insights into grassroots discourse, they often overlook official media narratives pivotal in shaping institutional and geopolitical agendas. Moreover, they seldom provide annotations for thematic framing, limiting their utility in studying state-aligned media strategies. As such, there remains a dearth of powerful resources to study longitudinal wartime narrative framing across multiple state-affiliated agencies. While two notable exceptions -- the VoynaSlov corpus~\cite{park2022challenges}, which captures Russian media and public responses, and the recent multidimensional analysis by \citet{ibrahim2025multi} -- offer the study of longitudinal framing (further discussed among related work in \S\hspace{1pt}\ref{sec:related-work}, they fall short of enabling contrastive analyses to examine how official news sources from different geopolitical regions differ in their narratives.
\section{Related Work}
\label{sec:related-work}
Research on wartime media coverage relies on datasets capturing narratives across national and institutional contexts, where competing claims and propaganda techniques pose unique challenges. Such resources are vital for training models to analyze geopolitical discourse and evaluating their ability to identify narrative strategies. As a reflection of diverse goals, existing resources vary in scope, modality, and annotations.

Datasets based on Twitter/X and Reddit are among the largest. \citet{chen2023tweets} aggregate over 570 million tweets in English, Russian, and Ukrainian to study information propagation and influence operations, while \citet{zhu2022reddit} provide 8 million posts correlating post categories with pro-Ukrainian stance. \citet{shevtsov2022twitter} introduce 57.3 million English tweets annotated for sentiment and hate speech, and the Invasion@Ukraine corpus provides 8.7 million English tweets for content moderation and conspiracy narrative analysis~\cite{pohl2023invasion}. The VoynaSlov collection~\cite{park2022challenges} bears similarities to our work through sentence-level media framing labels, though it focuses solely on Russian media.
\citet{alyukov2023wartime} introduce one of the first cross-platform social media datasets, integrating 1.7 million posts to enable the analysis of propaganda and public opinion.
Telegram-specific corpora include 4.1 million posts with pro/anti-Kremlin channel tags~\cite{bawa2025telegram} and 2.3 million posts with binary tags for human rights violations~\cite{nemkova2023detecting}.
\citet{hakimov2024} introduce a corpus of 1.5 million tweets in 60 languages with automated annotations  for sentiment, stance, and named entities. Their corpus is one of the widest multilingual social media datasets currently available.

News-based corpora are comparatively sparse. The RUWA corpus~\cite{khairova2024} provides 16{,}500 news articles using semantic similarity for misinformation detection, but excludes the overarching war theme and curates only specific event-based articles.
\textsc{dnipro} supersedes it by orders of magnitude in coverage, topics, multilinguality, temporality, and size.
A longitudinal media analysis corpus is provided by \citet{ibrahim2025multi}, with two years of news data, but its coverage is restricted to nations with auxiliary involvement in the Russo-Ukrainian war.

Annotation practices across existing resources differ widely: manual schemata for framing~\cite{park2022challenges}, channel stance~\cite{bawa2025telegram}, and human rights violations~\cite{nemkova2023detecting}; as well as automated large-scale labeling~\cite{pohl2023invasion, hakimov2024}.
Most resources rely on social media identifiers that degrade quickly, impeding reproduction.

\textsc{dnipro} addresses these gaps by providing longitudinally contiguous, multilingual news coverage from five primary geopolitical actors (Russia, Ukraine, U.S., U.K., and China), with comprehensive annotations for multiple downstream tasks.
Unlike social media-focused datasets optimized for volume and engagement patterns, \textsc{dnipro} enables systematic comparative analyses of state-aligned narrative framing across languages, nations, and outlets, over a duration of 2 years and 7 months, making it a unique and powerful resource for empirical studies on the creation and dissemination of competing wartime realities (\S\hspace{1pt}\ref{sec:applications}).

\section{The \textsc{dnipro} Corpus}
\label{sec:data-collection}
We present \textsc{dnipro}, a longitudinal corpus of 246{,}229 news articles spanning \textit{31 months} of the Russo-Ukrainian war (February 1, 2022 -- August 31, 2024).
Our corpus captures competing narratives from 11 outlets across five nation states (Russia, Ukraine, U.S., U.K., and China) and three languages (English, Russian, and Mandarin Chinese).
Its key descriptive statistics appear in \autoref{tab:descriptive-stats}.

Each article includes structured metadata following a unified schema (shown in \autoref{tab:metadata-schema}) designed to facilitate large-scale longitudinal and comparative analyses across sources, languages, and geopolitical contexts.
\textsc{dnipro} is constructed through a systematic multi-stage pipeline targeting news outlets and state-affiliated agencies selected on the basis of three criteria: (i) global influence and prominence within national media ecosystems, (ii) sustained coverage of the Russo-Ukrainian war, and (iii) discernible editorial positioning ranging from independent journalism to state-aligned messaging.
To the extent possible, access to full article archives was obtained through public RSS feeds, institutional subscriptions, or source-specific data collection pipelines.
However, not all media outlets provided consistent or unrestricted access to their archives, and some selections reflect practical constraints in addition to topical relevance.

We first detail our collection methodology (\S\ref{ssec:data-acquisition}), then describe the structure and access information for the data records (\S\ref{ssec:data-records}).

\begin{table}[!t]
\centering
{\small
\setlength{\tabcolsep}{3pt}
\begin{tabularx}{\linewidth}{@{}lX@{}}
\toprule
\multicolumn{2}{@{}l}{\textbf{Corpus Scale}}\smallskip\\
\hspace{5pt}Total articles & 246,229\\
\hspace{5pt}Avg. no. of articles/day & 261.1\\
\hspace{5pt}Title length (avg.) & 11.7 words$^*$ \\
\hspace{5pt}Article length (avg.) & 332.5 words$^*$\medskip\\
\multicolumn{2}{@{}l}{\textbf{Coverage}}\smallskip\\
\hspace{5pt}Temporal span & Feb 1, 2022 -- Aug 31, 2024\\
\hspace{5pt}No. of news sources & 11 sources\\
\hspace{5pt}Geopolitical entities & Russia, Ukraine, USA, UK, China\medskip\\
\multicolumn{2}{@{}l}{\textbf{Linguistic Composition}}\smallskip\\
\hspace{5pt}Languages covered & English, Russian, Mandarin Chinese\\
\hspace{5pt}Non-English articles & 61,637 (25.03\%) \medskip\\
\multicolumn{2}{@{}l}{\footnotesize $^*$\textit{After translation (of non-English sources to English)}.}\\
\bottomrule
\end{tabularx}
}
\caption{\small \textsc{dnipro} at a glance: scale, temporal and geographic coverage, and multilingual composition.}
\label{tab:descriptive-stats}
\end{table}

\subsection{Data Acquisition}
\label{ssec:data-acquisition}
\textsc{dnipro} comprises \textit{31 months} of continuous daily data collection (Feb. 1, 2022 -- Aug. 31, 2024).
To identify articles with substantive coverage of the Russo-Ukrainian conflict, we employ a tiered strategy.
For outlets with dedicated sections or labels related to the war, we collect all articles published under those categories.
Where available, we also extract articles tagged with conflict-specific metadata (e.g., ``Ukraine conflict'', ``special military operation'', ``Russia-Ukraine war'').
For sources lacking such taxonomies, we apply keyword-based filtering to isolate relevant content.
However, some sources impose significant restrictions on retrospective API-based keyword searches.
In these cases, we obtain the older articles directly through web browsers, operating within the scope of respective terms of service and copyright restrictions.
To ensure quality control, we manually validate stratified random samples across sources, languages, and time periods throughout the acquisition process.

{
\setlength{\tabcolsep}{3pt}
\begin{table}[!t]
\small
\centering
\begin{tabularx}{\linewidth}{@{}lX@{}}
\toprule
\textbf{Field} & \textbf{Description}\medskip\\
\multicolumn{2}{@{}l}{\textit{Identifiers \& Source}}\\
\hspace{5pt}\texttt{article\_id} & Unique UUID v4 identifier\\
\hspace{5pt}\texttt{url} & URL of the article\\
\hspace{5pt}\texttt{source\_name} & Normalized source name (e.g., \textit{Xinhua})\\
\hspace{5pt}\texttt{source\_country} & Country of origin (ISO 3166-1 alpha-3)\medskip\\
\multicolumn{2}{@{}l}{\textit{Content \& Language}}\\
\hspace{5pt}\texttt{title} & Article headline\\
\hspace{5pt}\texttt{article\_length} & Word count (post-processing)\\
\hspace{5pt}\texttt{language} & Article language (ISO 639-1)\medskip\\
\multicolumn{2}{@{}l}{\textit{Temporal Information}}\\
\hspace{5pt}\texttt{publication\_date} & Publication date (ISO 8601)\\% yyyy-mm-dd
\hspace{5pt}\texttt{collection\_date} & Date of URL access (ISO 8601)\\
\bottomrule
\end{tabularx}
\caption{\small Core metadata schema for \textsc{dnipro} articles.}
\label{tab:metadata-schema}
\end{table}
}
\textbf{English-language ``Western'' media} from the United States and the United Kingdom, namely \textit{The Financial Times} (FT) and \textit{Cable News Network} (CNN), were included as part of the Western bloc, reflecting the perspectives of countries with significant political, economic, and military involvement in the conflict. These outlets serve domestic and international audiences, and offer central platforms to help us understand how Western-aligned narratives are constructed and disseminated. No additional filtering of articles is done by search queries as these media houses offer dedicated sections for content related to the Russo-Ukrainian war. We collect a total of 5{,}760 and 4{,}218 articles from CNN and FT, respectively.

To ensure adequate representation of \textbf{Chinese state-aligned perspectives}, we collect articles from three leading outlets: \textit{Xinhua}, \textit{Global Times}, and \textit{China~Daily}.  
Xinhua publishes exclusively in Mandarin, whereas Global Times and China~Daily provide parallel English-language coverage.  
Articles are collected from Xinhua using a bilingual keyword list, designed to minimize the loss of coverage while retaining topical precision.
\begin{CJK}{UTF8}{gbsn}
    Mandarin Chinese queries comprised ``乌克兰'' (Ukraine), ``俄罗斯'' (Russia) and ``俄乌'' (Russia and Ukraine); 
\end{CJK}
English queries included ``Russia--Ukraine'', ``Russia'', ``Ukraine'', and ``Ukrainian''.  
These queries yield in 11{,}152 and 10{,}206 articles in Mandarin Chinese and English, respectively.

\begin{figure*}[!ht]
\centering
\begin{minipage}[c]{0.48\textwidth}
\centering
% [trim={left bottom right top},clip]
\includegraphics[trim={0 0 0 24pt}, clip, width=\linewidth]{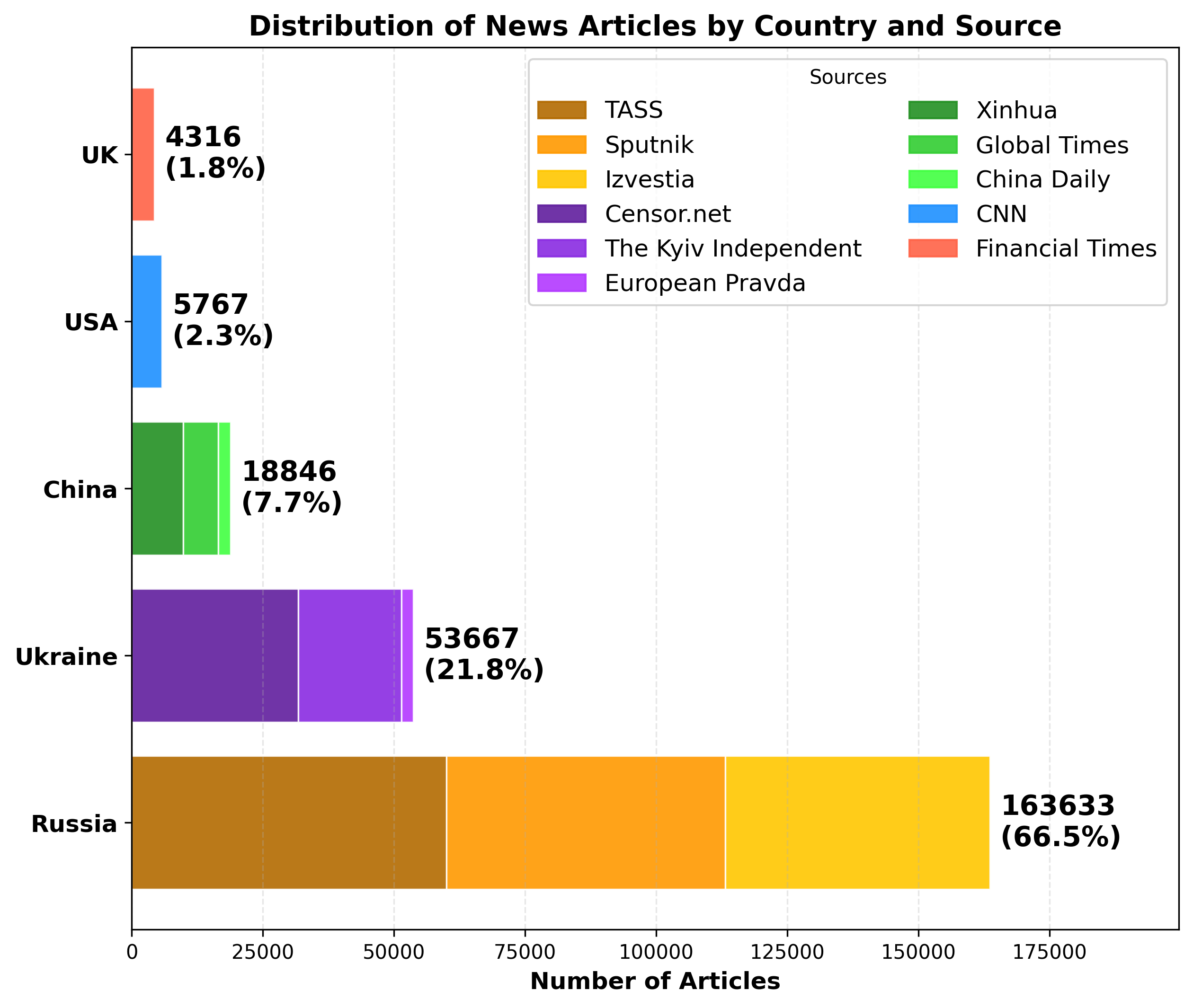}
\end{minipage}%
\hfill%
\begin{minipage}[c]{0.48\textwidth}
\centering
% [trim={left bottom right top},clip]
\includegraphics[trim={0 0 0 24pt}, clip, width=\linewidth]{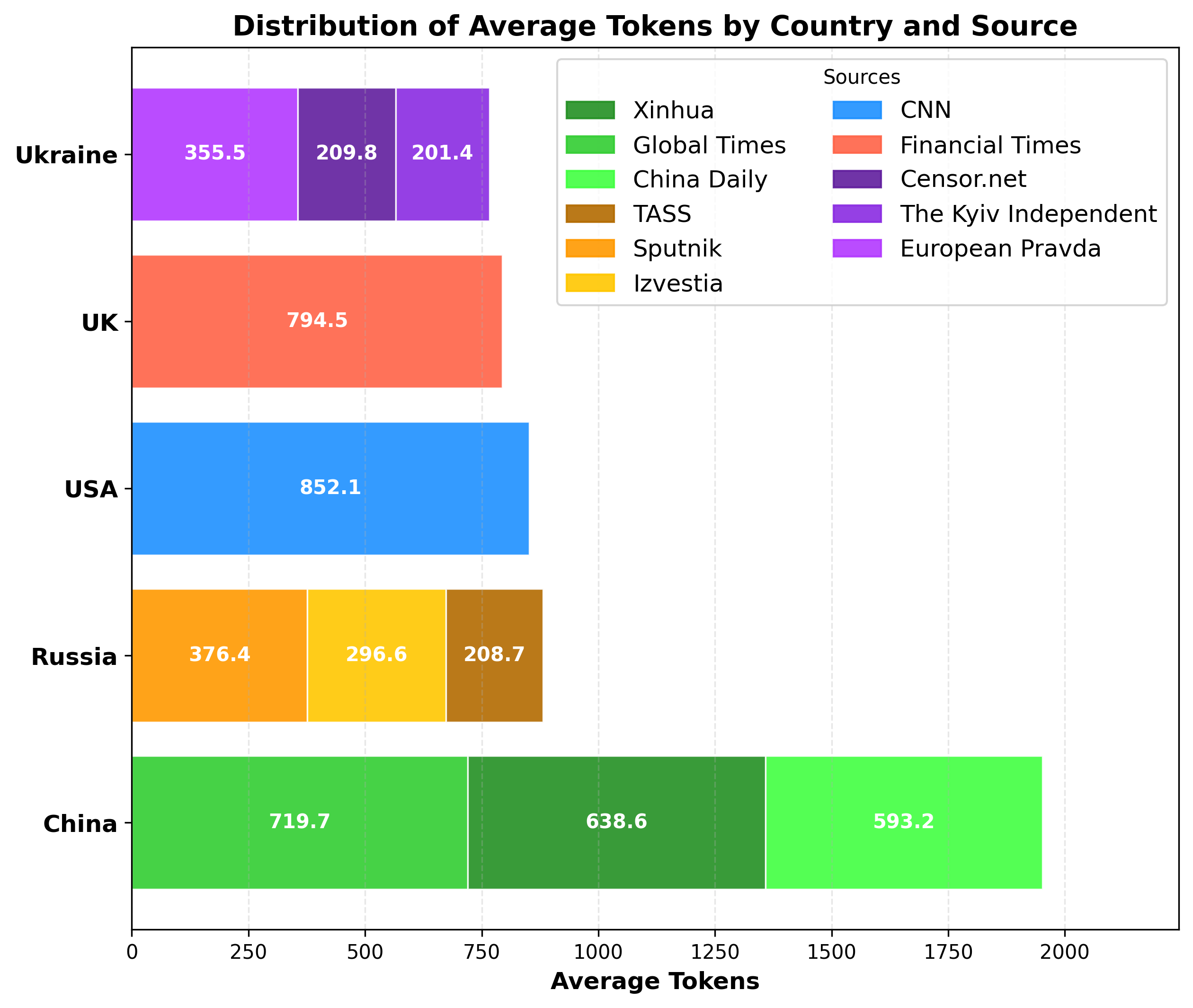}
\end{minipage}%
\caption{\small Composition of \textsc{dnipro}, across all sources, languages (English, Russian\textsuperscript{$\dagger$}, and Mandarin\textsuperscript{$\ddagger$}), and national affiliations: \textit{Censor.net}, \textit{The Kyiv Independent}, and \textit{European Pravda} from Ukraine; \textit{Sputnik}, \textit{TASS}, and \textit{Izvestia}\textsuperscript{$\dagger$} from Russia; \textit{CNN} from the USA; \textit{The Financial Times} from the UK; and \textit{China Daily}, \textit{Global Times}, and \textit{Xinhua}\textsuperscript{$\ddagger$} from China. Above, we see the distribution of articles (left) and the average word count of articles (right).}
\label{fig:distribution-by-country-and-source}
\end{figure*}

The \textbf{Ukrainian perspective} is represented by articles from \textit{Censor.net}, \textit{The Kyiv Independent}, and \textit{European Pravda}, all of which provide English-language versions.
While Censor.net influences Ukraine's domestic audience, the other two shape her global image.
%While the first two shape Ukraine’s global image, \textit{Censor.net} influences her domestic audience.
Their reporting styles vary---from Censor’s rapid updates to the in-depth opinions and interviews found in European Pravda.
As sources targeting the global Anglosphere, The Kyiv Independent and European Pravda offer sections dedicated to the Russo-Ukrainian War, labeled respectively with ``War'' and ``War with Russia''. The news reporting approach of Censor.net largely covers the war, evidenced by the majority fraction of Ukrainian war-related tags in their ``News'' section. As such, no keyword search queries are applied to any of the Ukrainian sources. The collection from these sources amounts to 19{,}706 articles from The Kyiv Independent, 2{,}212 from European Pravda, and 31{,}748 from Censor.net.

The \textbf{Russian media} sources in our dataset comprise English articles from \textit{TASS} and \textit{Sputnik}, and Russian-language articles from \textit{Izvestia}.
No keyword search queries are required, as all articles are collected from the dedicated section ``Special Operation of Russia in Ukraine''.
We collect 60{,}002 and 53{,}139 articles from TASS and Sputnik respectively, and 50{,}485 articles from Izvestia.

\autoref{fig:distribution-by-country-and-source} shows the distribution of articles and their average lengths across individual news sources and aggregated by geopolitical entities, providing an overview of the varying scales and styles of coverage across geopolitical perspectives.

\subsection{Data Records}
\label{ssec:data-records}
The \textsc{dnipro} corpus is available at no cost for non-commercial use under the \verb|CC-BY-NC 4.0| license,\footnote{The complete corpus is available on Zenodo: \href{https://doi.org/10.5281/zenodo.18470677}{doi.org/10.5281/zenodo.18470677}} and the corresponding processing scripts and annotation prompts are available on GitHub.\footnote{\href{https://github.com/dikshyam/dnipro-codebase}{github.com/dikshyam/dnipro-codebase}}
% It is structured for ease of use and distributed as a single, compressed \verb|tar.bz2| archive containing a primary metadata file, annotation files, and a documentation file.
It is structured for ease of use and distributed as separate downloadable files, comprising a primary metadata file, annotation files, and a documentation file. Due to copyright restrictions, we cannot directly include the texts. The provided URLs, however, serve as persistent identifiers enabling reproduction of \textsc{dnipro}, like most social media datasets. If an article becomes unavailable, the \texttt{collection\_date} and \texttt{url} fields enable retrieval via the \href{https://web.archive.org/}{Internet Archive's WayBack Machine}.\footnote{\href{https://web.archive.org/}{web.archive.org}}
\begin{enumerate}[nolistsep, noitemsep]
\item \textbf{\texttt{metadata.parquet}} contains the complete set of 246,229 metadata records as described in \autoref{tab:metadata-schema}. The Parquet format was chosen for its efficient compression and columnar storage, which enables fast querying and analysis using popular data science tools.
\item \textbf{\texttt{named-entities.parquet}} contains the named entity annotations for the entire corpus. Each record includes an \verb|article_id| (see \autoref{tab:metadata-schema}) and the list of named entities in that article.
%The annotations are done using SpaCy’s NER pipeline. This resource enables entity frequency analysis, and temporal tracking of key actors, organizations, locations, and events across the dataset.
\item \textbf{\texttt{topical-frames.parquet}} provides one row per \verb|article_id| with its corresponding \textbf{topical frames} anchored in categories proposed by \citet{habermas1991} (\autoref{tab:frame-definitions}).
%This schema is inspired by the Refugees and Migration Framing Vocabulary (RMFV; \citealp{yu_fliethmann_2022}). For each article we include \verb|labels| (the set of frames selected) and \verb|confidences| (frame-wise scores in $[0,1]$ indicating strength of evidence, where larger values reflect a stronger signal). Annotations are produced automatically with the \textsc{Phi\-4} 14B LLM; the exact prompt and decoding configuration appear in the appendix \ref{app:?}. 
\item \textbf{\texttt{sentiments.parquet}} provides sentiment scores for select entities and events across multiple news sources. For politically important and frequently occurring entities (\textit{e.g.}, ``Ukraine'', ``Russia'', ``Zelensky'', ``Kherson''), these scores are aggregated monthly. For entities tied to key events within this war (\textit{e.g.}, ``Bucha'' or ``Zaporizhzhia''), they are aggregated daily. The events used in this work are described in \S\hspace{1pt}\ref{sec:applications} (\autoref{tab:use-case-events}).
%Average Sentiment scores are computed monthly for high-frequency entities (Ukraine, Russia, Zelensky, Kherson) and daily for key war related events (Bucha, Kursk, Zaporizhzhia). This enables fine-grained temporal sentiment analysis and comparative narrative studies across countries and outlets.
\item \textbf{\texttt{README.md}} is a comprehensive description of the dataset, the schemata, the license terms, and a code of conduct for ethical use.
\end{enumerate}
We do not specify \verb|train|/\verb|test|/\verb|dev| splits for \textsc{dnipro}.
The corpus spans time, languages, and multiple sources; and supports diverse study designs (\textit{e.g.}, temporal evolution or development of an entity/topic/event; media source comparisons; transnational or cross-lingual reporting of events; or other types of aggregation).
A preset split would constrain the diversity of possible applications. We encourage task-appropriate partitions instead.
\section{Data Annotations}
\label{sec:data-annotations}
We now describe the enrichment of raw article metadata with annotations. Since copyright restrictions prevent the direct release of texts, we generate content-derived labels that form the dataset's core analytical value. The process involves:
\begin{itemize}[leftmargin=14pt, noitemsep, nolistsep]
\item metadata and text normalization;
\item translation of non-English text to English;
\item labeling named entities;
\item sentiment analysis;
\item stance analysis;
\item topical framing; and
\item rigorous quality control with human evaluation.
\end{itemize}

\subsection{Text Processing for Annotation}
\label{ssec:data-processing}
To prepare the corpus for annotation and analyses, we apply standardized processing steps to all data, across sources and languages. This ensures consistency across media sources with varying conventions in structure, formatting, and metadata. %Moreover, at this stage we remove non-content and duplicate content.

%\paragraph{Text and metadata normalization:}
Publication dates are parsed and normalized to ISO 8601 standard (\textsc{yyyy-mm-dd}), and multiple metadata fields---\textit{e.g.}, ``publication date'', ``published on''---are mapped to metadata fields in our unified schema (as shown in \autoref{tab:metadata-schema}). 
The article text content is stripped of boilerplate elements like location or date prefixes commonly found in news articles (\textit{e.g.}, ``KYIV, March 3 –''), inline footnotes, bylines, and repeated source disclaimers.
Duplicate articles are removed, to avoid redundancy in print syndication and multilingual reprints.
To support downstream linguistic analyses, the text is then converted to UTF-8 encoding, segmented into paragraphs, and split into sentences.
% using language-specific tokenization modules: ... for English, ... for Russian, and ... for Mandarin Chinese. \rb{specify the tools used for English, Russian, and Chinese}

{
\setlength{\tabcolsep}{2pt}
\begin{table}[t]
\centering
\small
\begin{tabularx}{\linewidth}{@{}p{1.6cm} X@{}}
\toprule
\textbf{Frame} & \textbf{Definition}\\
\midrule
\textsc{economy}\textsuperscript{\,1} & Macroeconomic issues and markets: sanctions/trade, inflation, growth, jobs, taxes, energy/commodity prices.\\
\textsc{identity}\textsuperscript{\,3} & Group identity and belonging: nationality, ethnicity, religion, language, or gender; discrimination or identity politics.\\
\textsc{morality}\textsuperscript{\,2} & Explicit ethical/value framing: moral duty, fairness, dignity, humanitarian arguments.\\
\textsc{legal}\textsuperscript{\,1} & Courts and legal processes: rulings, lawsuits, prosecutions, designations, warrants, legal rights or entitlements.\\
\textsc{policy}\textsuperscript{\,1} & Concrete government outlook or outreach, rules, bills, decrees, program design and details (what a policy does).\\
\textsc{politics}\textsuperscript{\,1} & Political proceedings and party competition: elections, horse-race strategy, reshuffles, coalition dynamics.\\
\textsc{public opinion}\textsuperscript{\,1} & Public attitudes and moods: polls, surveys, and protests as expressions of public sentiment and interest.\\
\textsc{security}\textsuperscript{\,1} & War/defense, violence/crime, policing, border enforcement, terrorism, cyberattacks.\\
\textsc{welfare}\textsuperscript{\,1} & Benefits and human services: healthcare, pensions, housing aid, unemployment, assistance to refugees or displaced people.\\
\bottomrule
\end{tabularx}
% \vspace{-2mm}
\caption{\small%
Topical frames in \textsc{dnipro}: an abridgment of the \textit{Refugees and Migration Framing} schema \cite{yu2022frame}, with its \textit{utilitarian}\textsuperscript{\,1}, \textit{moral-universal}\textsuperscript{\,2}, and \textit{identity-related}\textsuperscript{\,3} categories derived from the discourse ethics of \citet{habermas1991}.}
\label{tab:frame-definitions}
\end{table}
}

\subsection{Machine Translation}
\label{ssec:translation}
A quarter of \textsc{dnipro} consists of Mandarin Chinese and Russian texts, (\verb|zh| and \verb|ru|, respectively, in ISO 639-1). We translate these into English (\verb|en|) to facilitate downstream analyses. We choose three models based on recent comparative studies on neural machine translation by \citet{smirnov2022quantitative} and \citet{alemayehu2024error}: OPUS-MT~\cite{tiedemann2023opusmt}, M2M100~\cite{fan2020beyond}, and NLLB-200-Distilled~\cite{nllbteam2022}. Since there were no readily available explicit benchmarks for \verb|zh-to-en| translation of war or conflict narratives, we select a random set of ten articles from \textit{Xinhua}, evaluated by two native speakers of Mandarin Chinese for semantic fidelity and fluency.
% \begin{itemize*}
% \item[(i)] \verb|Helsinki-NLP/opus-mt-zh-en|~\cite{tiedemann2023opusmt},
% \item[(ii)] \verb|NLLB-200 Distilled 600M|~\cite{facebook2023nllb}, and
% \item[(iii)] \verb|M2M100-418M|~\cite{fan2020beyond}.
% \end{itemize*}
The evaluators worked independently to score the output of these models on 1-5 Likert scale for fidelity and fluency, and \verb|M2M100-418M| received the highest scores from both evaluators on both criteria. So, we employ this model for all \verb|zh-to-en| translations.
Russian articles from \textit{Izvestia} are similarly evaluated by native Russian speakers, and translated using the \verb|ru-to-en| path of \verb|M2M100-418M|.

\subsection{Annotation Methodology}
\label{ssec:annotation-methodology}
\noindent\textbf{Named Entity Recognition:}
To enable fine-grained analysis of key actors and locations, we perform named entity recognition (NER) across the entire corpus (after translation to English).
This annotation is carried out using spaCy~\cite{honnibal2020spacy} and its pretrained model \verb|en_core_web_sm|, identifying entities and their types according to the OntoNotes 5.0 taxonomy~\citeplanguageresource{ontonotes}.
This identifies 85{,}155 distinct named entity mentions, and provides the foundational layer for downstream tasks such as tracking the prominence of specific actors over time, temporal studies of media focus across entities and events, and entity-specific sentiment and stance analyses.

\noindent\textbf{Sentiment Analysis:}
To capture the sentiment surrounding the most prominent named entities, we provide \textit{sentence-level} sentiment annotation using a pretrained model: \texttt{sentiment-roberta-large-english}~\cite{hartmann2023}.
For each article, sentences containing such entities (\textit{e.g.}, \verb|Kursk|, \verb|Bucha|, \verb|Kherson|, \verb|Zelensky|, \verb|Ukraine|, \verb|Russia|)
are identified using spaCy, and passed to this classifier.
The model outputs are then mapped to continuous sentiment scores, which are then aggregated on a daily and monthly basis at both the source and country level using source–country relations.
These annotations thus provide time series data revealing how sentiment toward specific actors and events evolves across geopolitical perspectives.

\bgroup
\setlength{\tabcolsep}{1pt}
\begin{table}[!t]
\small
\centering
\begin{tabularx}{\linewidth}{@{}l r@{}}
\toprule
\textbf{Event (Date range)} & \textbf{\# articles}\\
\midrule
Bucha Massacre (Apr 3–9, 2022) & 273\\
Zaporizhzhia Nuclear Crisis (Mar 3–9, 2022) & 219\\
Mariupol Hospital Attack (Mar 9–15, 2022) & 138\\
Kursk Offensive (Aug 6–12, 2024) & 180\\
\bottomrule
\end{tabularx}
% \vspace{-3mm}
\caption{\small The four conflict events analyzed in our use case experiments, selected based on global coverage and the potential to generate narrative divergence.}
\label{tab:use-case-events}
\end{table}
\egroup

\vspace*{.5\baselineskip}
\noindent\textbf{Stance Detection:}
The dataset includes stance annotations obtained by few-shot instruction prompting with Phi-4 14B model ~\cite{phi4}, identifying the stance \textit{towards a specified target}.
The input comprises a sentence and the immediate context in its news article.
To Phi-4, we provide the task instruction with a few examples, along with (a) a passage for contextual information, (b) a focal sentence, and (c) the target. The model is prompted to output a compact, formatted response:
\bgroup
\setlength{\tabcolsep}{3pt}
{\small
\begin{myquote}
    \begin{tabular}{l p{.75\linewidth}}
    \texttt{Target:} & \texttt{<the identified target>}\\
    \texttt{Stance:} & \texttt{\textbf{favor}|\textbf{against}|\textbf{none}}\\
    \texttt{Rationale:} & \texttt{<a short justification grounded in the given text>}
    \end{tabular}
\end{myquote}
}
\egroup
\noindent\textbf{Topical Framing:}
For topical framing, we use the same few-shot prompting with Phi-4 14B (\citet{phi4}), where the input is an instruction describing the task, along with topical frame label definitions (shown in \autoref{tab:frame-definitions}) and 1-2 examples per label.
Then, for a given instance---comprising an article's title, subtitle (if present) and content---we prompt the model to generate a \verb|json| object containing the list of applicable topical frames and their corresponding confidence scores.
This supports analysis of how different sources emphasize humanitarian, military, diplomatic, or economic dimensions of the conflict.

\subsection{Human Evaluation}
\label{ssec:human-evaluation}
To assess annotation quality, we conduct rigorous human evaluation on both stance detection and topical framing.
Results demonstrate that automated labeling substantially agrees with expert judgment (\textsc{pabak} = 0.71 for topical framing, Fleiss' $\kappa$ = 0.643 for stance detection), validating their reliability for coarse-grain analyses while identifying specific categories that warrant careful interpretation.

\begin{table}[!t]
\centering
\small
\begin{tabularx}{\linewidth}{@{}X r r}
\toprule
\textbf{Annotator-pair}
    & \textbf{Agreement (\%)}
    & \textbf{Cohen's $\kappa$}\\
\midrule
Human annotators & 86.00\% & 0.770\\
Annotator\textsubscript{1} and \texttt{phi-4} & 72.68\% & 0.618\\
Annotator\textsubscript{2} and \texttt{phi-4} & 70.00\% & 0.580\smallskip\\
\multicolumn{3}{@{}l}{$\bullet\;\;$ Mean agreement: 76.23\%}\\
\multicolumn{3}{@{}l}{$\bullet\;\;$ \textbf{Fleiss' $\kappa$} across all three annotators: 0.643}\\
\multicolumn{3}{@{}l}{$\bullet\;\;$ Mean pairwise Cohen's $\kappa$: 0.656 (range: 0.58-0.77)}\\
\bottomrule
\end{tabularx}
% \vspace{-3mm}
\caption{\small%
Annotation agreement on stance labels: agreement percentage, pairwise Cohen's $\kappa$, and Fleiss' $\kappa$.}
\label{tab:kappa-results-stance}
\end{table}

\subsubsection{Stance Annotations}
\label{sssec:stance-annotations}
The stance annotation task identifies whether a sentence (with context) expressed a stance (\verb|favor| or \verb|against|) towards any entity, or if no stance was present (\verb|none|).
Two human annotators with native proficiency in English annotated a sample of 150 instances.
We evaluate their inter-annotator agreement, and then also with \verb|phi-4| as the LLM annotator.
Results in \autoref{tab:kappa-results-stance} present the pairwise and overall inter-annotator agreements, demonstrating strong human inter-annotator agreement ($\kappa$ = 0.77) and moderate LLM-human agreements ($\kappa$ = 0.62 and 0.58), showing consistent but less-than-human reliability.
The overall Fleiss' $\kappa$ = 0.643 indicates that the stance annotations are substantially reliable for large-scale analysis.

While the LLM shows a consistent gap with human judgment, this is expected for nuanced pragmatic language tasks like stance detection.
The annotations remain valuable for trend analysis and comparative studies across sources---such as identifying systematic differences in how Russian vs. Ukrainian sources frame political leadership or specific events---while nuanced high-stakes stance analyses may benefit from human review.

\bgroup
\setlength{\tabcolsep}{2.46pt}
\begin{table}[!t]
\centering
\small
\begin{tabularx}{\linewidth}{@{}l r r @{\hspace{4pt}} r r@{}}
\toprule
\textbf{Frame}
    & \multicolumn{2}{c}{\textbf{\textsc{pabak}}}
    & \multicolumn{2}{c}{\textbf{Confidence}}\\
\midrule
    & \multicolumn{1}{c}{Human} & \multicolumn{1}{c}{All} & \multicolumn{1}{c}{Human} & \multicolumn{1}{c}{Phi-4}\medskip\\
\multicolumn{5}{@{}l}{\textbf{High reliability (\textsc{pabak} > 0.8)}}\\
\hspace{11pt}Public Opinion & 0.87 (93.4\%) & 0.865 & 0.80 & 0.89\\
\hspace{11pt}Legal & 0.85 (92.4\%) & 0.783 & 0.74 & 0.91\\
\hspace{11pt}Welfare & 0.81 (90.4\%) & 0.783 & 0.80 & 0.87\medskip\\
\multicolumn{5}{@{}l}{\textbf{Substantial reliability (\textsc{pabak} 0.69 - 0.8)}}\\
\hspace{11pt}Morality & 0.75 (87.3\%) & 0.750 & 0.79 & 0.91\\
\hspace{11pt}Identity & 0.74 (86.8\%) & 0.783 & 0.79 & 0.92\\
\hspace{11pt}Economy & 0.70 (84.8\%) & 0.736 & 0.84 & 0.93\\
\hspace{11pt}Security & 0.69 (84.3\%) & 0.587 & 0.85 & 0.93\medskip\\
\multicolumn{5}{@{}l}{\textbf{Moderate reliability (\textsc{pabak} < 0.69)}}\\
\hspace{11pt}Policy & 0.55 (77.7\%) & 0.526 & 0.78 & 0.89\\
\hspace{11pt}Politics & 0.48 (74.1\%) & 0.425 & 0.79 & 0.92\medskip\\
\textbf{Average} & 0.71 (85.7\%) & 0.693 & 0.80 & 0.91\\
\bottomrule
\end{tabularx}
% \caption{Inter-annotator agreement (\textsc{pabak} and percentage agreement) among humans, and model-human agreement, along with frame-wise confidence scores for topical frame annotations.}
% \vspace{-3mm}
\caption{\small%
Human inter-annotator agreement scores (\textsc{pabak} and percentage), human-model agreement (\textsc{pabak}), and annotator confidence scores (\texttt{[0-1]}) across all nine topical frames introduced in \autoref{tab:frame-definitions}.}
\label{tab:kappa-results-framing}
\end{table}
\egroup

\subsubsection{Topical Frames}
\label{sssec:topical-frames}
Two expert annotators independently label a stratified random sample of 200 articles, providing nine binary labels with corresponding [0,1] confidence scores for each article.
Significant class imbalance is a common challenge in multi-label annotation, and the inter-annotator agreement must consider the high \textit{base rate agreement} where annotators agree on a label being \textit{absent} from an article.
The average agreement on \textit{label absence}, over all nine topical frames, accounts for 65.82\% of our evaluation sample, with the positive class (for any label) representing a minority of instances.
Due to this prevalence bias, traditional measures like Cohen's or Fleiss' $\kappa$ are overly pessimistic reliability measures for our data.
For a more accurate assessment that accounts for this imbalance, we report the \textit{prevalence-adjusted bias-adjusted kappa}, or \textsc{pabak}~\cite{byrt1993bias}, which is especially crucial for evaluation of consensus on the rarer labels.

The \textbf{inter-annotator agreement} results, measured by \textsc{pabak}, are presented in \autoref{tab:kappa-results-framing} and demonstrate substantial agreement on average (0.71, with 85.7\% agreement). Three topical frame categories --- \verb|Welfare|, \verb|Public Opinion|, and \verb|Legal| --- reveal near-perfect agreement ($>$ 0.8). Overall, seven of the nine labels achieve \textsc{pabak} scores $\geq$ 0.7, indicating a consistent and shared understanding of the annotation framework.
The two remaining classes, \verb|Policy| and \verb|Politics|, have moderate agreement, as they represent more abstract and often overlapping conceptual domains.

Further, we examine the correlation between the average confidence scores reported by the human experts and those reported by \verb|phi-4|.
The framing categories are complex and sometimes overlapping, resulting in a weak positive correlation $\rho$ = 0.37.
Notably, \verb|phi-4| exhibits systematically higher confidence (avg. 0.91) than human annotators (avg. 0.80) across all categories, regardless of the extent of agreement.
This suggests that users should not interpret \verb|phi-4|'s self-reported confidence scores as reliable indicators of annotation quality.

\vspace*{.5\baselineskip}
\noindent\textbf{Implications for Dataset Use:}
The strong agreement validates the overall reliability of our annotation schema. However, we recommend that users consider the following nuances in their studies:
\begin{itemize}[leftmargin=16pt, nolistsep, noitemsep]
\item[1)] Frames with \textit{high/substantial reliability} (\textsc{pabak} > 0.69) are suitable for fine-grained analysis such as tracking shifts in \verb|Welfare| framing after major tragedies, or comparing \verb|Security| or \verb|Legal| discourse across state-aligned media.
\item[2)] The \textit{moderately reliable} frames (\textsc{pabak} < 0.69) --- \verb|Policy| and \verb|Politics| --- yield lower agreement due to their inherent conceptual overlap. These frame labels may be best suited for coarse-grained topic modeling or as supplementary features, with the understanding that they represent broader discourse categories that often intersect. As such, their unification may provide more reliable signals.
\item[3)] The expert-annotated sample of 200 documents, stratified across sources, languages, and time periods, provides a completely validated high-quality \textit{gold-standard benchmark}. Our stratified sampling ensures it is suitable for evaluating the performance of topical frame classification systems on real-world data on contentious wartime narratives.
\end{itemize}
Together, these evaluations demonstrate that \textsc{dnipro}'s annotations achieve a balance of scale and quality rare in wartime media corpora: automated methods enable comprehensive coverage, while stratified human validation ensures researchers can make informed decisions regarding labels, as per their task-specific goals. The annotation guidelines and task-specific prompts are provided in Appendix A and C, respectively.
\section{Use Case Experiments}
\label{sec:applications}
To demonstrate the analytical potential of \textsc{dnipro}, we present four use-case experiments exploiting its unique multi-perspective structure.
These are not exhaustive analyses, but proof-of-concept demonstrations showing how the corpus enables diverse research in computational journalism, political communication, discourse analysis, and information warfare studies.
Each application highlights specific affordances of the corpus while suggesting broader research directions.

\begin{figure}[!t]
\centering
\includegraphics[width=\linewidth]{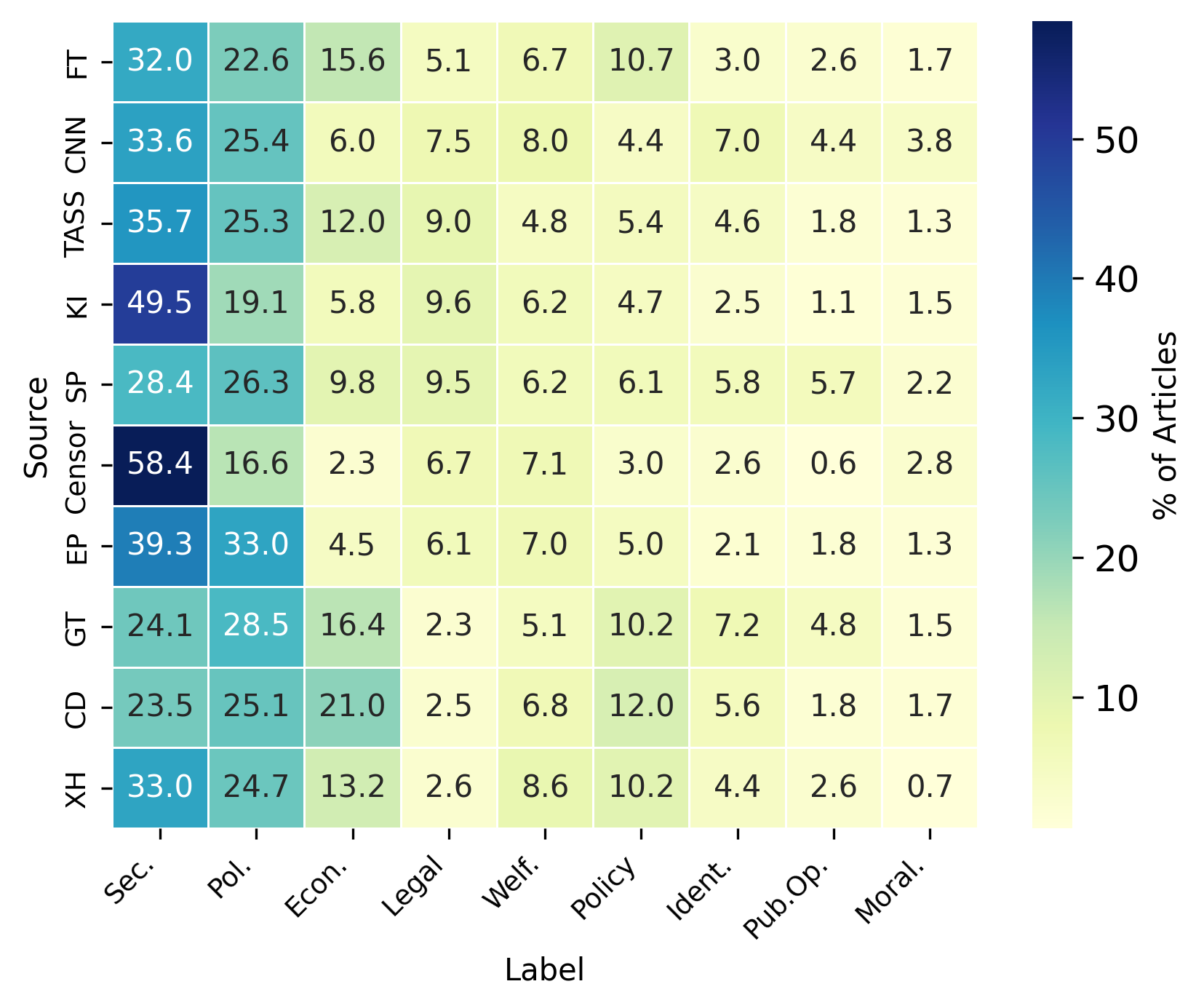}
% \vspace{-4mm}
\caption{\small%
Topical framing across news outlets (for each frame, \% of articles with confidence > 0.5): \texttt{Security} dominates all outlets, while \texttt{Economy} is most prominent in Chinese media and \textit{Financial Times} (\textsc{ft}). Other abbr.: \textit{Kyiv Independent} (\textsc{ki}), \textit{Sputnik} (\textsc{sp}), \textit{European Pravda} (\textsc{ep}), \textit{Global Times} (\textsc{gt}), \textit{China Daily} (\textsc{cd}), \textit{Xinhua} (\textsc{xh}).}
\label{fig:topical_framing}
\end{figure}

\vspace*{.5\baselineskip}
\noindent\textbf{1. A Thematic Study of Conflict Events:}
Our analysis focuses on the four major conflict events shown in \autoref{tab:use-case-events}, which received substantial global coverage, generated clear geopolitical divisions, and invited transnational comparisons.
For each event, we filter articles using specific location identifiers (\textit{e.g.}, \verb|Bucha|) to ensure relevance.

% To examine how different themes are prioritized when covering conflict events, we analyze the topical emphasis patterns across outlets.
% We identify which concepts, actors, and subtopics receive attention in coverage of major events, revealing differential emphasis on aspects such as \textit{historical context}, \textit{humanitarian concerns}, or \textit{institutional responses}.
We examine how conflict events are framed by analyzing patterns of thematic salience across news outlets.
By identifying which concepts, actors, and subtopics are underscored in coverage, we reveal how outlets differentially emphasize aspects such as \textit{historical context}, \textit{humanitarian concerns}, or \textit{institutional responses}.

% Using GPT-3.5~\cite{ye2023gpt} with structured prompts, we tag articles from weekly time windows surrounding each major event to identify key concepts, actors, and subtopics.
% For the events described in \autoref{tab:use-case-events}, we obtain 810 articles, with 10.5 such tags (avg.) per article.
% We aggregate these tags by event and country to find the distinct thematic emphasis patterns across media outlets.

Word clouds visualizing these thematic patterns for each conflict event and outlet country are provided in Appendix B. Our prelimnary analysis shows that Russian and Chinese outlets consistently emphasize institutional terminology (\textit{special military operation}), security concerns (\textit{nuclear weapons}), and counter-narratives (\textit{media manipulation}, \textit{ceasefire}), reflecting strategic adversarial framing.
Ukrainian outlets predominantly highlight humanitarian dimensions (\textit{civilian casualties}) with frequent references to specific conflict locations.
Media from the USA and the UK balance humanitarian topics with institutional responses (\textit{sanctions}, \textit{international criminal court}) and policy discussions (\textit{no fly zone}).

Thus, \textsc{dnipro} captures differences in thematic salience and agenda-setting across geopolitical actors narrating the same underlying events.

\vspace*{.5\baselineskip}
\noindent\textbf{2. Topical Framing:}
We conduct a corpus-wide analysis using the schema introduced in \autoref{tab:frame-definitions}.
Article-level topical frames provide a comparable map of what each media outlet emphasizes, revealing systematic priorities that contextualize downstream analyses such as stance, sentiment, or discourse.
Articles are annotated with \verb|phi-4| using a few-shot instruction prompt, producing multi-label assignments along with confidence scores (as described in \S\hspace{1pt}\ref{sssec:topical-frames}).

\autoref{fig:topical_framing} shows that \verb|Security| dominates across all outlets, with the highest shares in Ukrainian media, followed by Russian state media and Western outlets like CNN.
The second-most prevalent category is \verb|Politics|, elevated in Russian and Chinese media. It is substantially present in other outlets too.
\verb|Economy|
%-- capturing sanctions, energy, and market impacts --
is most visible in Chinese media and, naturally, The Financial Times, while \verb|Policy| appears steadily across outlets, with higher emphasis in Chinese media and the UK.
Relative to others, Russian media leans more on \verb|Legal| topics.
% \textsc{Public Opinion}, \textsc{Welfare}, and \textsc{Identity} each constitute low single-digit shares.
Notably, discussions of \verb|Morality| are extremely rare despite the focal topic of this dataset being an extended war, indicating that operational, strategic, and material considerations far outweigh explicit ethical rhetoric in war coverage.

\begin{figure}[!t]
\centering
\includegraphics[width=\linewidth]{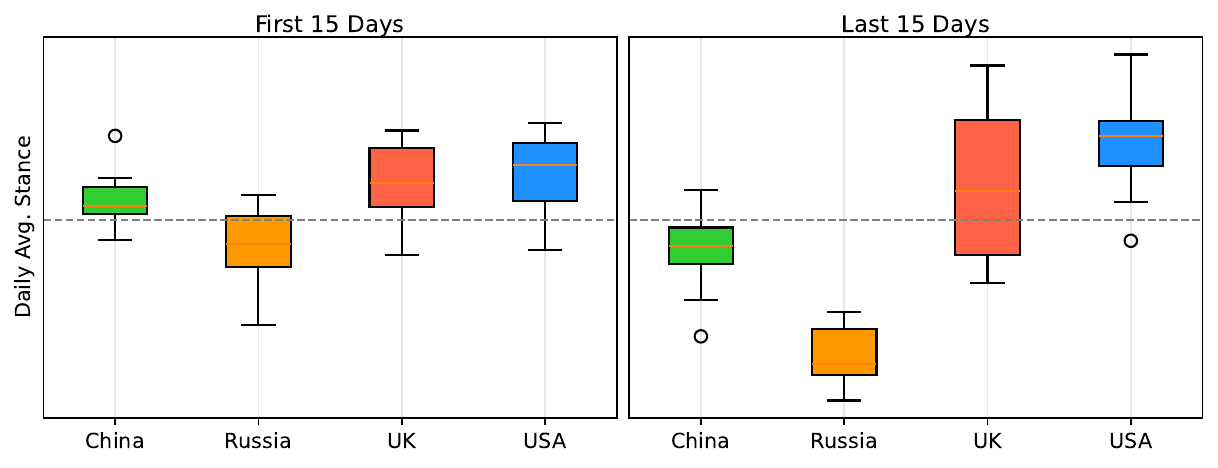}
% \vspace{-2mm}
\caption{\small Daily average stance scores toward \texttt{Ukraine}, grouped by nation, comparing the first 15 days of the war (left) with the last 15 days of data collection (right). Note the dramatic intensification of Russian negativity and American support, increased variability in the UK, and the gentle negative shift in Chinese media.}
\label{fig:stance-boxplot}
\end{figure}

\vspace*{.5\baselineskip}
\noindent\textbf{3. Stance Analysis:}
We compare stance toward \verb|Ukraine| between the war's first 15 days and the final 15 days of data collection (\autoref{fig:stance-boxplot}), revealing striking temporal shifts.
Chinese media shifted from mildly pro-Ukrainian to mildly oppositional, likely reflecting China's deepening strategic alignment with Russia.
Russian coverage moved from a mildly negative stance to overwhelmingly hostile -- possibly correlating with battlefield changes and political rhetoric. The US coverage became even more supportive, while the UK exhibited greater variability, possibly due to the transition from early solidarity to war fatigue, domestic economic pressures and/or leadership changes.

The temporal dimension of \textsc{dnipro} proves crucial for these insights: China's pivot, Russian and American intensification in opposite directions, and growing ambivalence in the UK.

\vspace*{.5\baselineskip}
\noindent\textbf{4. Divergent Narratives:}
To investigate how media outlets construct and contest narratives~\cite{gamson1992media, takeshita1997exploring}, we analyze claims to identify conflicting representations of key events. Building on prior work in conflict journalism~\cite{entman2010framing, thussu2017china}, this analysis unearths how opposing perspectives are embedded in individual claims.

Using weekly data for each event, we extract sentence-level claims and categorize them into six types, motivated by epistemology \cite{chafe1986evidentiality}, subjectivity and stance analysis \cite{wiebe2005annotating, ferreira2016emergent}, and speech acts \cite{austin1962things, searle1969speech}:
\begin{itemize}[leftmargin=4pt, nolistsep, noitemsep]
\item[]\textbf{\texttt{factual}}: direct statements of observable or confirmed events;
\item[]\textbf{\texttt{attributed}}: claims attributed to specific actors;
\item[]\textbf{\texttt{disputed}}: explicit denials or rebuttals; 
\item[]\textbf{\texttt{reported}}: indirect accounts from third-parties;
\item[]\textbf{\texttt{command/policy}}: official statements or directives;
\item[]\textbf{\texttt{analytical/assessment}}: interpretive or evaluative judgments by experts or institutions.
\end{itemize}
With GPT-3.5 plus human-in-the-loop validation, we label claims via structured prompts, and group semantically related claims through agglomerative clustering on Sentence Transformer representations~\cite{reimers2019sentence}. 

\noindent
Our key findings demonstrate \textsc{dnipro}'s utility in the study of \textit{information warfare}, \textit{perception}, and \textit{distortion}~\cite{galtung1965structure} --- not as explicit propaganda, but as systematic divergence of constructed realities across geopolitical divides:
\begin{itemize}[leftmargin=*, nolistsep, noitemsep]
\item \textbf{Contradictions are usually implicit:}
Contradictions predominantly emerge across geopolitical divides as competing interpretations.
Outlets rarely offer explicit rebuttals, instead they construct incompatible narratives through different topical frames, omissions, or counter-attributions and selective emphasis (\autoref{tab:contradiction-pairs}).
\item \textbf{Alignment with geopolitical blocs:} 
Entailment occurs primarily within aligned media ecosystems: Ukrainian and Western outlets reinforce each other through overlapping \texttt{factual}, \texttt{reported}, or \texttt{command/policy} claims, while Russian media maintains internal consistency through \texttt{attributed} and \texttt{analytical} claims citing official statements.
\item \textbf{Framing substitutes for rebuttal:} When covering the same event (e.g., an airstrike), Russian media might characterize it as precision targeting of military assets (\texttt{factual/attributed}), while Western media offer contrary eyewitness accounts (\texttt{disputed/reported}).
\end{itemize}

{
\begin{table}[!t]
\small
\centering
\setlength{\tabcolsep}{3pt}
\begin{tabularx}{\linewidth}{@{}l X@{}}
\toprule
\multicolumn{2}{c}{\textbf{Bucha Massacre}}\\
\midrule
\textsc{ru} & \yh{Russia had testimony from residents} of Bucha and Mariupol, who reported brutal crimes, such as torture, \yh{committed by Ukrainian nationalists}. \textit{(TASS)}\\
\textsc{ua} & \rh{All eyewitnesses noted} the systematic nature of the killings in Bucha \rh{by the Russian army}. \textit{(Censor)}\\
\midrule
\multicolumn{2}{c}{\textbf{Sumy Oblast Attacks}}\\
\midrule
\textsc{ru} & \yh{Russian forces jointly with border guards repelled these attacks} and hit enemy positions near the border and in the Sumy Region. \textit{(TASS)}\\
\textsc{ua} & \rh{Russian forces attacked six border areas and settlements of Sumy Oblast} on June 7, firing 14 times and causing at least 54 explosions \ldots \textit{(Kyiv Independent)}\\
\bottomrule
\end{tabularx}
\caption{\small Competing realities narrated by Russian and Ukrainian media: Russian sources attribute atrocities to Ukrainian forces (yellow), while Ukrainian sources blame Russian forces (red).
Narratives are opposed in attribution and framing, absent direct refutation or negation.}
\label{tab:contradiction-pairs}
\end{table}
}
\section{Conclusion}
\label{sec:conclusion}
We present \textsc{dnipro}, a longitudinal corpus of 246K news articles capturing competing geopolitical narratives of the Russo-Ukrainian war across 31 months (Feb 2022 -- Aug 2024).
Unlike prior corpora focused on social media volume or single-language coverage, \textsc{dnipro} enables comparative analysis across nation-states through
structured, multilingual news data with human-evaluated annotations for stance, sentiment, and topical framing.

Our experiments demonstrate \textsc{dnipro}'s capacity to support diverse analytical approaches and reveal systematic patterns, from thematic emphasis patterns and temporal stance tracking to fine-grained claim-level contradiction analysis.
This makes \textsc{dnipro} uniquely suited for studying information ecosystems where competing narratives coexist without direct engagement, and \textit{contradictions are constructed implicitly}: media outlets build parallel realities through divergent attribution and selective emphasis rather than direct refutation.

This has methodological implications: traditional approaches assume explicit falsehoods or rebuttals~\cite{scantamburlo2026scoping, guo2022survey}, but competing narratives operate through \textit{incompatible framing}. Thus, data-driven research needs new methods that map how attribution choices and temporal patterns construct divergent realities across information ecosystems.

\textsc{dnipro} enables scalability in several research directions: 
\begin{itemize*}%[leftmargin=*, noitemsep]
\item[(a)] longitudinal narrative evolution (e.g., what are the predictors of geopolitical realignments);
\item[(b)] cross-lingual information flow (e.g., which media outlets serve as bridges, and which ones serve as echo chambers);
\item[(c)] event-driven polarization: (e.g., can we predict which events trigger lasting narrative shifts); and 
\item[(d)] computational detection of incompatible narrative frames, in the absence of explicit disagreement markers.
\end{itemize*}

As international conflicts unfold across the fragmented media landscape, understanding these dynamics becomes essential for researchers, policymakers, journalists, and citizens. 
\textsc{dnipro} is publicly available under CC-BY-NC 4.0:
\begin{itemize}[leftmargin=*, nolistsep, noitemsep]
\item[] \textbf{Dataset:} \href{https://doi.org/10.5281/zenodo.18470677}{doi.org/10.5281/zenodo.18470677}
\item[] \textbf{Code:} \href{https://github.com/dikshyam/dnipro-codebase}{github.com/dikshyam/dnipro-codebase}
\end{itemize}
We invite researchers to utilize \textsc{dnipro}, and develop new methods for detecting, tracking, and understanding narrative divergence: work that bridges computational analysis with urgent questions about truth, persuasion, and power in the digital age.

% \noindent
% \textbf{Dataset:} \href{https://doi.org/10.5281/zenodo.18470677}{doi.org/10.5281/zenodo.18470677}\\
% \textbf{Annotation Scripts:} \href{https://github.com/dikshyam/dnipro-codebase}{github.com/dikshyam/dnipro-codebase}

% \noindent
% We invite researchers to develop new methods for detecting, tracking, and understanding narrative divergence—work that bridges computational analysis with urgent questions about truth, persuasion, and power.

% As international conflicts unfold across the fragmented media landscape, understanding these dynamics becomes essential for researchers, policymakers, journalists, and citizens. \textsc{dnipro} is publicly available under a CC-BY-NC 4.0 license at \href{https://doi.org/10.5281/zenodo.18470677}{doi.org/10.5281/zenodo.18470677}, with annotation scripts hosted at \href{https://github.com/dikshyam/dnipro-codebase}{github.com/dikshyam/dnipro-codebase}.
% We invite researchers to utilize \textsc{dnipro}, and develop new methods for detecting, tracking, and understanding narrative divergence --- work that bridges computational analysis with urgent questions about truth, persuasion, and power in the digital age.
% limitations and ethics don't count towards the page limit %
% doesn't count towards page limit
\section{Limitations}
\noindent\textbf{Temporal scope:}
\textsc{dnipro} covers February 2022 through August 2024, but the Russo-Ukrainian War has not concluded.
Thus, significant developments after this period --- including shifts in foreign policy, territorial control, or international alliances --- are not captured.
Researchers studying more recent events should consider this temporal boundary when interpreting findings, or extend our corpus to include later developments.

\noindent\textbf{Geographic coverage:}
Our selection stresses on geopolitical actors directly involved (Russia and Ukraine) and influential despite no direct engagement in the war (the United States, the United Kingdom, and China).
NATO representation, however, is limited to two of its thirty-two member states.
Perspectives from several influential NATO states (e.g., Germany and its state-own media, Deutsche Welle) are absent.
Regional actors and nations with complex positions may also be included in the analyses we have underscored in our work.
Such actors include Hungary and Romania, both of which border Ukraine; Belarus, also bordering Ukraine, but currently allied with the Russian state; T\"urkiye, a powerful regional actor, especially in the Black Sea; Georgia, a NATO contender that continues to maintain reasonably strong relations with Russia; and India, which maintains nuanced multi-alignments.
The inclusion of these actors may reveal more sophisticated patterns of alignment and hedging than the polarized perspectives currently represented.

\noindent\textbf{Annotation reliability:}
While our human evaluations demonstrate substantial inter-annotator agreement (\S\hspace{1pt}\ref{ssec:human-evaluation}), automated ground-truth annotations are never perfect.
This is not a limitation for qualitative analyses, mixed-method approaches, or empirical analyses for coarse-grained signals; but research requiring finer labels for specific nuanced tasks may benefit from additional human validation of their task-relevant subsets of \textsc{dnipro}.

\noindent\textbf{Scope of use-case experiments:}
Our exploratory analyses demonstrate analytical possibilities rather than provide definitive empirical claims.
These demonstrations are not limitations of the corpus.
Rather, they serve as proof-of-concept for the types of research \textsc{dnipro} enables: longitudinal stance tracking, framing comparison across media outlets and nation states, and contradiction detection.
But we emphasize that these are not exhaustive treatments of these topics.
Each use case is illustrative, and intended for deeper investigations.

\noindent\textbf{Translation and linguistic nuances:}
Non-English articles are machine-translated to English for cross-lingual analysis.
Even as this enables systematic comparisons, translations may lose culturally specific rhetoric or connotations that carry meaning in the original language.
Researchers with relevant language expertise may benefit from analyzing the untranslated Russian and Chinese Mandarin articles for such discursive phenomena.

%%% older draft 
% For the chosen news sources, a wider variety of publications representing NATO nations would be beneficial.
% Currently, only two of the thirty-two NATO nations are represented (the United Kingdom and the United States of America).
% There are other media sources like DW in Germany, which are highly referenced globally, so they are an influential factor.
% Additionally, representation of more countries with ambiguous involvement in the war would add value to a truly global version of the dataset.
% Potentially interesting countries would be Hungary and Romania, both of which border Ukraine; Belarus, also bordering Ukraine, but currently allied with the Russian state; T\"urkiye, a powerful regional actor, especially in the Black Sea; and Georgia, a NATO contender that continues to maintain reasonably strong relations with Russia.

% Another limiting facet of \textsc{dnipro} is the imperfect ground-truth annotations for stance, and other discursive phenomena. While this is not a limitation for qualitative analyses, mixed-method approaches, or broad empirical analyses, it does hinder finer nuanced large-scale learning directly from the corpus. Our exploratory analyses, while illustrative, are not exhaustive and serve primarily as demonstrations of the dataset's potential rather than to make definitive empirical claims.
% doesn't count towards 8 page limit
\section{Ethical Considerations}
\textsc{dnipro} documents an ongoing conflict with significant human costs and geopolitical implications. We acknowledge several potential ethical dimensions:
\begin{itemize}[leftmargin=-0.6pt, nolistsep]
\item[]\textit{Dual-use considerations.}
       This corpus enables research into how narratives shape perception, with applications in computational journalism, political communication, and information literacy.
       However, the same analytical methods could be misused to refine propaganda or manipulate opinion.
       We urge researchers to consider the societal implications of their work and to strive for transparency, cross-cultural understanding, and informed discourse.
\item[]\textit{Source inclusion and representation.}
       \textsc{dnipro} inherently reflects the biases present in global media coverage.
       The inclusion of opposing narratives---including state-controlled media---is a deliberate choice, to empower comparative analyses.
       \textbf{Inclusion does not imply endorsement of content}.
       Some sources may contain disinformation or inflammatory rhetoric.
       We include them because the research possibilitied outlined earlier require access to the full spectrum of discourse, not sanitized versions.
       Researchers should contextualize sources appropriately using the provided metadata (media outlet ownership, editorial stance, etc.).
       We strongly recommend any derived datasets or findings to retain all the provided metadata, to ensure that articles can be interpreted within their proper editorial and temporal contexts.
\item[]\textit{Sensitive content.}
       This corpus pertains to an armed conflict. Many articles may describe violence and other distressing aspects of armed conflicts.
       Researchers working with this data should be aware of this, particularly when conducting qualitative analysis or annotation tasks.
       We recommend appropriate support structures for research teams.
\item[]\textit{Data provenance and access,}
       All content was gathered from publicly available news sources without breaching access restrictions or terms of service.
       We provide URLs rather than full text to respect intellectual property rights and enable verification, and do not include any copyright-protected content or any private communications.
       The CC-BY-NC 4.0 license permits research and educational use while prohibiting commercial exploitation.
% \item[]\textit{Contextualization and metadata.}
%        We provide robust metadata (publication date, source country, outlet ownership) to ensure articles can be interpreted within their proper editorial and temporal contexts rather than in isolation.
%        We strongly recommend that any derived datasets or findings retain this contextual information.
\end{itemize}

\noindent
We designed \textsc{dnipro} to support research that illuminates how information ecosystems function during contested events --- work that ultimately serves democratic accountability, informed citizenship, and conflict resolution.

\noindent
We earnestly welcome community input on additional safeguards or ethical considerations.

\section{Bibliographical References}
\bibliographystyle{lrec2026-natbib}
\bibliography{references}

@article{galtung1965structure,
  title={{The Structure of Foreign News: The Presentation of the Congo, Cuba and Cyprus Crises in Four Norwegian Newspapers}},
  author={Galtung, Johan and Ruge, Mari Holmboe},
  journal={Journal of Peace Research},
  volume={2},
  number={1},
  pages={64--91},
  year={1965},
  publisher={Sage Publications},
  doi = {10.1177/002234336500200104}
}

@book{chafe1986evidentiality,
  editor={Chafe, Wallace and Johanna Nichols},
  title={{Evidentiality: The Linguistic Coding of Epistemology}},
  publisher={Ablex},
  year={1986},
  address={Norwood, NJ},
  series={Advances in Discourse Processes},
  volume={20}
}

@inproceedings{wiebe2005annotating,
  author = {Wiebe, Janyce and Wilson, Theresa and Cardie, Claire},
  title = {{Annotating Expressions of Opinions and Emotions in Language}},
  booktitle = {Language Resources and Evaluation},
  year = {2005},
  publisher = {Springer},
  pages = {165--210},
  volume = {39},
  doi = {10.1007/s10579-005-7880-9}
}

@inproceedings{ferreira2016emergent,
    title = "{E}mergent: a novel data-set for stance classification",
    author = "Ferreira, William  and
      Vlachos, Andreas",
    editor = "Knight, Kevin  and
      Nenkova, Ani  and
      Rambow, Owen",
    booktitle = "Proceedings of the 2016 Conference of the North {A}merican Chapter of the Association for Computational Linguistics: Human Language Technologies",
    year = "2016",
    address = "San Diego, California",
    publisher = "Association for Computational Linguistics",
    doi = "10.18653/v1/N16-1138",
    pages = "1163--1168"
}

@book{searle1969speech,
  title={{Speech Acts: An Essay in the Philosophy of Language}},
  author={Searle, John R.},
  year={1969},
  doi={10.1017/CBO9781139173438},
  publisher={Cambridge University Press}
}

@book{austin1962things,
  title={{How to Do Things with Words}},
  author={Austin, John Langshaw},
  year={1962},
  publisher={Oxford University Press}
}

@book{habermas1991,
  author    = {Habermas, J{\"u}rgen},
  title     = {{Erl{\"a}uterungen zur Diskursethik}},
  address   = {Frankfurt am Main},
  publisher = {Suhrkamp},
  series    = {Suhrkamp Taschenbuch Wissenschaft},
  number    = {975},
  year      = {1991}
}

@article{guo2022survey,
  author = {Guo, Zhijiang and Schlichtkrull, Michael and Vlachos, Andreas},
  title = {{A Survey on Automated Fact-Checking}},
  journal = {Transactions of the Association for Computational Linguistics},
  volume = {10},
  pages = {178--206},
  year = {2022},
  month = {02},
  doi = {10.1162/tacl_a_00454}
}

@article{scantamburlo2026scoping,
  author = {Scantamburlo, Teresa and Zollo, Fabiana},
  title = {{A Scoping Review of Misinformation Research: Definitions, Conceptualizations, and Measurement Approaches}},
  year = {2026},
  publisher = {Association for Computing Machinery},
  address = {New York, NY, USA},
  doi = {10.1145/3796546},
  journal = {ACM Trans. Web}
}

@article{gamson1992media,
  title     = {{Media Images and the Social Construction of Reality}},
  author    = "Gamson, William A. and Croteau, David and Hoynes, William and Sasson, Theodore",
  journal   = "Annual Review of Sociology",
  publisher = "Annual Reviews",
  volume    =  18,
  number    =  1,
  pages     = "373--393",
  year      =  "1992",
  doi = {10.1146/annurev.so.18.080192.002105} 
}

@incollection{takeshita1997exploring,
  author    = {Takeshita, Toshio},
  title     = {{Exploring the Media's Roles in Defining Reality: From Issue-Agenda Setting to Attribute-Agenda Setting}},
  booktitle = {Communication and Democracy: Exploring the Intellectual Frontiers in Agenda-Setting Theory},
  editor    = {McCombs, Maxwell and Shaw, Donald L. and Weaver, David},
  pages     = {15--27},
  year      = {1997},
  publisher = {Lawrence Erlbaum Associates},
  address   = {New York},
  url ={https://www.taylorfrancis.com/chapters/edit/10.4324/9780203810880-3}
}

@phdthesis{werd2018thesis,
  author = "Werd, P. G. de",
  title = {{Critical Intelligence: Analysis by Contrasting Narratives}},
  school = "Universiteit Utrecht",
  year = "2018",
  url = {http://hdl.handle.net/1874/373430},
  type = "Doctoral Thesis"
}

@article{saletta2020,
  title     = {{The role of narrative in collaborative reasoning and intelligence analysis: A case study}},
  author    = "Saletta, Morgan and Kruger, Ariel and Primoratz, Tamar and Barnett, Ashley and van Gelder, Tim and Horn, Robert E.",
  journal   = "PLoS One",
  publisher = "Public Library of Science (PLoS)",
  volume    =  15,
  number    =  1,
  pages     = "e0226981",
  year      =  2020,
  doi = {10.1371/journal.pone.0226981}
}

@inproceedings{alyukov2023wartime,
  title = {{Wartime Media Monitor ({W}ar{MM}-2022): A Study of Information Manipulation on {R}ussian Social Media during the {R}ussia-{U}kraine War}},
  author = "Alyukov, Maxim and Kunilovskaya, Maria and Semenov, Andrei",
  editor = "Degaetano-Ortlieb, Stefania and Kazantseva, Anna and Reiter, Nils and Szpakowicz, Stan",
  booktitle = "Proceedings of the 7th Joint SIGHUM Workshop on Computational Linguistics for Cultural Heritage, Social Sciences, Humanities and Literature",
  month = {May},
  year = "2023",
  address = "Dubrovnik, Croatia",
  publisher = "Association for Computational Linguistics",
  doi = "10.18653/v1/2023.latechclfl-1.17",
  pages = "152--161"
}

@article{bawa2025telegram,
  title={{Telegram as a Battlefield: Kremlin-Related Communications During the Russia-Ukraine Conflict}},
  volume={19},
  doi={10.1609/icwsm.v19i1.35939},
  number={1},
  journal={Proceedings of the International AAAI Conference on Web and Social Media},
  author={Bawa, Apaar and Kursuncu, Ugur and Achilov, Dilshod and Shalin, Valerie L. and Agarwal, Nitin and Akbas, Esra},
  year={2025},
  month={June},
  pages={2361-2370}
}

@article{byrt1993bias,
  title = {Bias, prevalence and kappa},
  journal = {Journal of Clinical Epidemiology},
  volume = {46},
  number = {5},
  pages = {423-429},
  year = {1993},
  doi = {10.1016/0895-4356(93)90018-V},
  author = {Ted Byrt and Janet Bishop and John B. Carlin}
}

@article{chen2023tweets,
  title={{Tweets in Time of Conflict: A Public Dataset Tracking the Twitter Discourse on the War between Ukraine and Russia}},
  volume={17},
  doi={10.1609/icwsm.v17i1.22208},
  number={1},
  journal={Proceedings of the International AAAI Conference on Web and Social Media},
  author={Chen, Emily and Ferrara, Emilio},
  year={2023},
  month={June},
  pages={1006-1013}
}

@inproceedings{hakimov2024,
  author = {Hakimov, Sherzod and Cheema, Gullal S.},
  title = {{Unveiling Global Narratives: A Multilingual Twitter Dataset of News Media on the Russo-Ukrainian Conflict}},
  year = {2024},
  publisher = {Association for Computing Machinery},
  address = {New York, NY, USA},
  doi = {10.1145/3652583.3657622},
  booktitle = {Proceedings of the 2024 International Conference on Multimedia Retrieval},
  pages = {1160–1164},
  location = {Phuket, Thailand},
  series = {ICMR '24}
}

@inproceedings{khairova2024,
  title={{Unsupervised approach for misinformation detection in Russia-Ukraine war news}},
  author={Nina Khairova and Andrea Galassi and Fabrizio Lo Scudo and Bogdan Ivasiuk and Ivan Redozub},
  booktitle={Proceedings of the 8th International Conference on Computational Linguistics and Intelligent Systems. Volume IV: Computational Linguistics Workshop},
  pages={21--36},
  year={2024},
  url = {https://ceur-ws.org/Vol-3722/paper3.pdf},
  publisher={CEUR Workshop Proceedings}
}

@misc{nemkova2023detecting,
  title={{Detecting Human Rights Violations on Social Media during Russia-Ukraine War}}, 
  author={Poli Nemkova and Solomon Ubani and Suleyman Olcay Polat and Nayeon Kim and Rodney D. Nielsen},
  year={2023},
  eprint={2306.05370},
  archivePrefix={arXiv},
  primaryClass={cs.CY},
  doi={10.48550/arXiv.2306.05370},
  url={https://arxiv.org/abs/2306.05370}
}

@article{pohl2023invasion,
  title={{Invasion@Ukraine: Providing and Describing a Twitter Streaming Dataset That Captures the Outbreak of War between Russia and Ukraine in 2022}},
  volume={17}, 
  doi={10.1609/icwsm.v17i1.22217},
  number={1},
  journal={Proceedings of the International AAAI Conference on Web and Social Media},
  author={Susanne Pohl, Janina and Markmann, Simon and Assenmacher, Dennis and Grimme, Christian},
  year={2023},
  month={June},
  pages={1093-1101}
}

@article{gamson1989media,
  author = {Gamson, William A. and Modigliani, Andre},
  title = {{Media Discourse and Public Opinion on Nuclear Power: A Constructionist Approach}},
  journal = {American Journal of Sociology},
  year = {1989},
  volume = {95},
  number = {1},
  pages = {1--37},
  doi = {10.1086/229213}
}

@book{fairclough2010critical,
  author = {Fairclough, Norman},
  title = {{Critical Discourse Analysis: The Critical Study of Language}},
  edition = {2},
  publisher = {Routledge},
  year = {2010},
  doi = {10.4324/9781315834368}
}

@article{entman1993framing,
  author = {Entman, Robert M.},
  title = {{Framing: Toward Clarification of a Fractured Paradigm}},
  journal = {Journal of Communication},
  year = {1993},
  volume = {43},
  number = {4},
  pages = {51--58},
  doi = {10.1111/j.1460-2466.1993.tb01304.x}
}

@article{hartmann2023,
  title = {{More than a Feeling: Accuracy and Application of Sentiment Analysis}},
  author = {Jochen Hartmann and Mark Heitmann and Christian Siebert and Christina Schamp},
  journal = {International Journal of Research in Marketing},
  volume = {40},
  number = {1},
  pages = {75--87},
  year = {2023},
  doi = {10.1016/j.ijresmar.2022.05.005}
}

@misc{phi4,
  author = {Microsoft},
  title = {{Phi-4 Technical Report}},
  year = {2024},
  url = {https://www.microsoft.com/en-us/research/publication/phi-4-technical-report/},
  note = {Accessed: Oct 17, 2025}
}

@book{thussu2017china,
  title = {{China's Media Go Global}},
  editor = {Thussu, Daya Kishan and De Burgh, Hugo and Shi, Anbin},
  year={2017},
  publisher={Routledge},
  address={London},
  doi={10.4324/9781315619668}
}

@article{zhang2016china,
  author = {Zhang, Xiaoling and Wasserman, Herman and Mano, Winston},
  title = {{China’s Expanding Influence in Africa: Projection, Perception and Prospects in Southern African Countries}},
  journal = {Communicatio},
  year = {2016},
  volume = {42},
  number = {1},
  pages = {1--22},
  doi = {10.1080/02500167.2016.1143853}
}

@article{roman2017information,
  author = {Roman, Nataliya and Wanta, Wayne and Buniyak, Iuliia},
  title = {{Information wars: Eastern Ukraine military conflict coverage in the Russian, Ukrainian and U.S. newscasts}},
  journal = {International Communication Gazette},
  year = {2017},
  volume = {79},
  number = {4},
  pages = {357--378},
  doi = {10.1177/1748048516682138}
}

@article{nygren2016journalism,
  author = {Nygren, Gunnar and Glowacki, Michal and H\"{o}k, J\"{o}ran and Kiria, Ilya and Orlova, Dariya and Taradai, Daria},
  title = {{Journalism in the Crossfire: Media Coverage of the War in Ukraine in 2014}},
  journal = {Journalism Studies},
  year = {2016},
  volume = {19},
  number = {7},
  pages = {1059--1078},
  doi = {10.1080/1461670X.2016.1251332}
}

@article{ibrahim2025multi,
  author = {Ibrahim, Majd and Wang, Bang and Xu, Minghua and Xu, Hang},
  title = {{A multidimensional analysis of media framing in the Russia-Ukraine war}},
  journal = {Journal of Computational Social Science},
  year = {2025},
  volume = {8},
  number = {34},
  doi = {10.1007/s42001-025-00363-1}
}

@inproceedings{park2022challenges,
  title = {{Challenges and Opportunities in Information Manipulation Detection: An Examination of Wartime Russian Media}},
  author = "Park, Chan Young and Mendelsohn, Julia and Field, Anjalie and Tsvetkov, Yulia",
  editor = "Goldberg, Yoav and Kozareva, Zornitsa and Zhang, Yue",
  booktitle = "Findings of the Association for Computational Linguistics: EMNLP 2022",
  year = "2022",
  publisher = "Association for Computational Linguistics",
  doi = "10.18653/v1/2022.findings-emnlp.382",
  pages = "5209--5235"
}

@article{shevtsov2022twitter,
  title={{Twitter Dataset on the Russo-Ukrainian War}},
  author={Alexander Shevtsov and Christos Tzagkarakis and Despoina Antonakaki and Polyvios Pratikakis and Sotiris Ioannidis},
  year={2022},
  volume={2204.08530},
  journal={arXiv},
  doi={10.48550/arXiv.2204.08530}
}

@article{zhu2022reddit,
  title={{A Reddit Dataset for the Russo-Ukrainian Conflict in 2022}},
  author={Yiming Zhu and Ehsan-ul Haq and Lik-Hang Lee and Gareth Tyson and Pan Hui},
  year={2022},
  volume={2206.05107},
  journal={arXiv},
  doi={10.48550/arXiv.2206.05107}
}

@mastersthesis{silvertsen2024projecting,
  author = {Silvertsen, Julia},
  title = {{Projecting Geopolitical Power and Acting as One: The EU’s Foreign Policy in the Russian-Ukrainian Conflict [2014 and 2022-23]}},
  school = {Norwegian University of Science and Technology},
  year = {2024},
  address={Trondheim, Norway},
  url={https://hdl.handle.net/11250/3142338}
}

@article{dervis2023transformation,
  title={{Transformation of Geopolitical Perceptions in the Russian-Ukrainian War: Impact on Regional Relations in the Future}},
  volume={1},
  doi={10.57125/FS.2023.03.20.02},
  number={1},
  journal={Futurity of Social Sciences},
  author={Dervi\c{s}, Leyla},
  year={2023},
  month={Mar},
  pages={21–34}
}

@misc{watling2023preliminary,
  author={Watling, Jack and Danylyuk, Oleksandr V. and Reynolds, Nick},
  title={{Preliminary Lessons from Russia’s Unconventional Operations During the Russo-Ukrainian War, February 2022–February 2023}},
  howpublished={Royal United Services Institute (RUSI) for Defence and Security Studies},
  year={2023},
  url={https://www.rusi.org/explore-our-research/publications/special-resources/preliminary-lessons-russias-unconventional-operations-during-russo-ukrainian-war-february-2022},
  note = {Date: Mar 29, 2023; Retrieved: Nov 28, 2024}
}

@article{bradshaw2024strategic,
  author = {Bradshaw, Samantha and Elswah, Mona and Haque, Monzima and Quelle, Dorian},
  title = {{Strategic Storytelling: Russian State-Backed Media Coverage of the Ukraine War}},
  journal = {International Journal of Public Opinion Research},
  volume = {36},
  number = {3},
  pages = {edae028},
  year = {2024},
  month = {July},
  doi = {10.1093/ijpor/edae028}
}

@article{johais2024unleash,
  author = {Eva Johais and Mareike Meis},
  title = {{`Unleash the hounds!': NAFO's memetic war narrative on the Russo-Ukrainian conflict}},
  journal = {Critical Studies on Security},
  pages = {1--13},
  year = {2024},
  volume = {13},
  number = {1},
  publisher = {Routledge},
  doi = {10.1080/21624887.2024.2395658}
}

@misc{russia2024ban,
  author={AFP},
  title={{Russia Bans Media Outlets From Using Words `War', `Invasion'}},
  howpublished={The Moscow Times},
  year = {2022},
  url={https://www.themoscowtimes.com/2022/02/26/russia-bans-media-outlets-from-using-words-war-invasion-a76605},
  note={Date: Feb 6, 2022; Retrieved: Nov 29, 2024}
}

@misc{onemillion,
  author={Pancevski, Bojan},
  title={{One Million Are Now Dead or Injured in the Russia-Ukraine War}},
  howpublished={The Wall Street Journal},
  year = {2024},
  url={https://www.wsj.com/world/one-million-are-now-dead-or-injured-in-the-russia-ukraine-war-b09d04e5},
  note={Date: Sep 17, 2024; Retrieved: Feb 3, 2025}
}

@article{tiedemann2023opusmt,
  author={Tiedemann, J{\"o}rg and Aulamo, Mikko and Bakshandaeva, Daria and Boggia, Michele and Gr{\"o}nroos, Stig-Arne and Nieminen, Tommi and Raganato, Alessandro and Scherrer, Yves and Vazquez, Raul and Virpioja, Sami},
  title  = {{Democratizing Neural Machine Translation with OPUS--MT}},
  journal= {Language Resources and Evaluation},
  year   = {2023},
  volume = {58},
  pages  = {713--755},
  doi    = {10.1007/s10579-023-09704-w},
  publisher={Springer Nature},
}

@misc{nllbteam2022,
  title={{No Language Left Behind: Scaling Human-Centered Machine Translation}}, 
  author={NLLB~Team},
  year={2022},
  eprint={2207.04672},
  archivePrefix={arXiv},
  primaryClass={cs.CL},
  url={https://arxiv.org/abs/2207.04672}, 
}

@article{fan2020beyond,
  author  = {Angela Fan and Shruti Bhosale and Holger Schwenk and Zhiyi Ma and Ahmed El-Kishky and Siddharth Goyal and Mandeep Baines and Onur Celebi and Guillaume Wenzek and Vishrav Chaudhary and Naman Goyal and Tom Birch and Vitaliy Liptchinsky and Sergey Edunov and Michael Auli and Armand Joulin},
  title   = {{Beyond English-Centric Multilingual Machine Translation}},
  journal = {Journal of Machine Learning Research},
  year    = {2021},
  volume  = {22},
  number  = {107},
  pages   = {1--48},
  url     = {http://jmlr.org/papers/v22/20-1307.html}
}

@inproceedings{alemayehu2024error,
    title = "Error Analysis of Multilingual Language Models in Machine Translation: A Case Study of {E}nglish-{A}mharic Translation",
    author = "Alemayehu, Hizkiel Mitiku  and
      Zahera, Hamada M  and
      Ngonga Ngomo, Axel-Cyrille",
    editor = "Al-Onaizan, Yaser  and
      Bansal, Mohit  and
      Chen, Yun-Nung",
    booktitle = "Proceedings of the 2024 Conference on Empirical Methods in Natural Language Processing",
    year = "2024",
    address = "Miami, Florida, USA",
    publisher = "Association for Computational Linguistics",
    doi = "10.18653/v1/2024.emnlp-main.1102",
    pages = "19758--19768"
}

@misc{akash2025largelanguagemodelsaddress,
      title={{Can Large Language Models Address Open-Target Stance Detection?}}, 
      author={Abu Ubaida Akash and Ahmed Fahmy and Amine Trabelsi},
      year={2025},
      eprint={2409.00222},
      archivePrefix={arXiv},
      primaryClass={cs.CL},
      url={https://arxiv.org/abs/2409.00222}, 
}

@article{entman2010framing,
  title={{Media framing biases and political power: Explaining slant in news of Campaign 2008}},
  author={Entman, Robert M.},
  journal={Journalism},
  volume={11},
  number={4},
  pages={389--408},
  year={2010},
  publisher={Sage Publications},
  doi={10.1177/1464884910367587}
}

@misc{honnibal2020spacy,
  author  = {Honnibal, Matthew and Montani, Ines and Van Landeghem, Sofie and Boyd, Adriane},
  title   = {{spaCy: Industrial-strength Natural Language Processing in Python}},
  year    = {2020},
  doi     = {10.5281/zenodo.1212303}
}

@inproceedings{reimers2019sentence,
    title = {{Sentence-BERT: Sentence Embeddings using Siamese BERT-Networks}},
    author = "Reimers, Nils  and
      Gurevych, Iryna",
    editor = "Inui, Kentaro  and
      Jiang, Jing  and
      Ng, Vincent  and
      Wan, Xiaojun",
    booktitle = "Proceedings of the 2019 Conference on Empirical Methods in Natural Language Processing and the 9th International Joint Conference on Natural Language Processing (EMNLP-IJCNLP)",
    year = "2019",
    publisher = "Association for Computational Linguistics",
    doi = "10.18653/v1/D19-1410",
    pages = "3982--3992"
}

@article{yu2022frame, 
    title={{Frame Detection in German Political Discourses: How Far Can We Go Without Large-Scale Manual Corpus Annotation?}}, 
    volume={35}, 
    doi={10.21248/jlcl.35.2022.227}, 
    number={2}, 
    journal={Journal for Language Technology and Computational Linguistics}, 
    author={Yu, Qi and Fliethmann, Anselm}, 
    year={2022}, 
    month={July}, 
    pages={15–31}
}

@inproceedings{smirnov2022quantitative,
  title={{Quantitative Comparison of Translation by Transformers-Based Neural Network Models}},
  author={Alexander V. Smirnov and Nikolay Teslya and Nikolay Shilov and Diethard Frank and Elena Minina and Martin Kovacs},
  booktitle={International Conference on Enterprise Information Systems},
  year={2022},
  doi={10.1007/978-3-031-39386-0_8}
}

@misc{openai2023gpt35,
  author = {{OpenAI}},
  title = {GPT-3.5 language model},
  year = {2023},
  howpublished = {\url{https://developers.openai.com/api/docs/models/gpt-3.5-turbo}}
}

@misc{brown2020languagemodelsfewshotlearners,
      title={Language Models are Few-Shot Learners}, 
      author={Tom B. Brown and Benjamin Mann and Nick Ryder and Melanie Subbiah and Jared Kaplan and Prafulla Dhariwal and Arvind Neelakantan and Pranav Shyam and Girish Sastry and Amanda Askell and Sandhini Agarwal and Ariel Herbert-Voss and Gretchen Krueger and Tom Henighan and Rewon Child and Aditya Ramesh and Daniel M. Ziegler and Jeffrey Wu and Clemens Winter and Christopher Hesse and Mark Chen and Eric Sigler and Mateusz Litwin and Scott Gray and Benjamin Chess and Jack Clark and Christopher Berner and Sam McCandlish and Alec Radford and Ilya Sutskever and Dario Amodei},
      year={2020},
      eprint={2005.14165},
      archivePrefix={arXiv},
      primaryClass={cs.CL},
      url={https://arxiv.org/abs/2005.14165}, 
}

\section{Language Resource References}
\bibliographystylelanguageresource{lrec2026-natbib}
\bibliographylanguageresource{resources}

\newpage
% RB:
% Rewriting the appendices.
% Context --> then the models --> and then the annotation types with explanation
% First explain what the prompt is for, only then provide the detailed prompt
% \appendix

\section*{Appendix A. Annotation Methodology}
\label{app-a:annotation-methodology}

This appendix provides complete prompt specifications and implementation details for the large-scale automated annotations described in \S\hspace{1pt}\ref{ssec:annotation-methodology}.
All prompts follow structured output formats to enable reliable parsing and post-processing.

\vspace*{.5\baselineskip}
\noindent\textbf{Models and APIs:}
We used the following language models for automated annotations:
\begin{itemize}[leftmargin=*, nolistsep, noitemsep]
\item \verb|microsoft/phi-4| (14B parameters)~\cite{phi4} for stance detection and topical framing;
\item a fine-tuned RoBERTa-large model by \citet{hartmann2023}\footnote{\texttt{siebert/sentiment-roberta-large-english} (Hugging Face).} for sentiment analysis; and
\item GPT-3.5 Turbo, via the OpenAI API (\citet{openai2023gpt35, brown2020languagemodelsfewshotlearners}), for thematic tagging and claim typology.
\end{itemize}

\vspace{\baselineskip}
\noindent
\textbf{Stance detection} proceeds in two stages:
\begin{itemize*}
\item[(1)] \textit{target generation} identifies stance-worthy entities or events in a sentence, and then
\item[(2)] \textit{stance classification} determines the expressed attitude toward each target.
\end{itemize*}
This decomposition improves accuracy by separating entity recognition from sentiment interpretation~\cite{akash2025largelanguagemodelsaddress}.
We apply stance detection to articles from: (a) the first and last 15 days of data collection, and (b) four major conflict events listed in \autoref{tab:use-case-events}: the Bucha massacre, the Kursk offensive, Mariupol hospital attack, and Zaporizhzhia nuclear crisis.

\vspace{.5\baselineskip}
\noindent\textit{Target Generation.}
The model identifies all stance-worthy targets -- actors, organizations, or events that can be supported or opposed -- mentioned in a sentence. The prompt template for target generation is shown next:

%[caption={\small Few-shot prompt for target generation}]
\begin{promptbox}
You are performing @\textbf{Target Detection for Stance Classification}@.

Task:
1. Read the sentence carefully.
2. Identify ALL the @\mbox{\textbf{stance-worthy targets}}@
   (entities or events) mentioned in the
   sentence.
3. A stance-worthy target can be:
   - An @\textbf{actor}@ (government, organization,
     leader) whose responsibilities or
     actions can be supported or opposed.
   - An @\textbf{event/claim}@ that people can express
     favor, opposition, or neutrality toward.
4. Return @\mbox{\textbf{multiple targets}}@ if present,
   separated by commas.
5. If no stance-worthy targets are found,
   return exactly: "Targets: NONE".
6. Output only the targets, do not repeat
   the sentence.
Examples:
1) @\mbox{\textbf{Input:}}@
   Sentence: "Ukrainian forces downed 22 of
   the 27 Shahed-type drones launched by
   Russia overnight on June 5, the Air Force
   said in its morning update."
   @\mbox{\textbf{Output:}}@
   Targets: Ukrainian forces, Russia

2) @\mbox{\textbf{Input:}}@
   Sentence: "Russian Defense Ministry said
   it also shot down Ukraine's Western-
   provided missiles."
   @\mbox{\textbf{Output:}}@
   Targets: Russian Defense Ministry, Ukraine,            Russia

Now analyze the following sentence:
   Sentence: {sentence}
   Output format:
   Targets: <comma-separated targets or NONE>
\end{promptbox}

\vspace{.5\baselineskip}
\noindent\textit{Stance Classification.}
For each detected target, the model classifies the stance expressed in the sentence using the surrounding passage as context, thereby resolving pronouns and implicit references.

\begin{promptbox}
You are performing @\mbox{\textbf{Stance Classification}}@, with context.

Task:
1. Read the full passage carefully.
2. Focus only on the highlighted sentence.
3. Given the detected target, classify the
   stance expressed in the specified sentence
   @\mbox{\textbf{towards that target}}@.
4. Additionally, provide a short reasoning
   (1-2 sentences) explaining why the stance
   is FAVOR, AGAINST, or NONE.

Labels:
  - FAVOR   = supportive or positive stance
  - AGAINST = negative stance
  - NONE    = neutral, descriptive, or
              unclear stance

Important:
  - Use the passage to resolve ambiguous
    references.
  - If the target is NONE, stance will also
    be NONE.
  - Keep the reasoning factual and grounded
    in the text.

---

### Examples:
1) @\mbox{\textbf{Input:}}@
     Passage: Ukrainian forces downed ...
     Sentence: "Ukrainian forces downed 22 of
     the 27 Shahed-type drones launched by
     Russia overnight on June 5, the Air
     Force said in its morning update."
     Target: Ukrainian forces
   @\mbox{\textbf{Output:}}@
     Stance: FAVOR
     Reason: The sentence highlights a
     successful action by Ukrainian forces,
     emphasizing their effectiveness in
     defense. It frames them positively by
     reporting their achievement in downing
     most of the drones.

---

Passage: {passage}
Specified Sentence: {sentence}
Target: {target}

@\mbox{\textbf{Output format:}}@
  Stance: <FAVOR / AGAINST / NONE>
  Reason: <brief reason for the stance>
\end{promptbox}

\vspace{.5\baselineskip}
\noindent
\textbf{Sentiment analysis} is performed at sentence-level, for key entities like \verb|Russia|, \verb|Ukraine|, \verb|Zelensky|, \verb|Bucha|, \verb|Kursk|, etc.
For each entity, we
\begin{itemize*}
\item[(a)] extract all sentences mentioning the entity using the English-language model of spaCy~\cite{honnibal2020spacy};
\item[(b)] pass each sentence to the fine-tuned RoBERTa-large model~\cite{hartmann2023};
\item[(c)] map model outputs to continuous scores in \verb|[-1, 1]|; and
\item[(d)] aggregate the scores daily \textit{and} monthly, by source \textit{and} country.
\end{itemize*}

\begin{promptbox}
Identify the sentiment (positive, negative, or neutral) towards {entity} in this context:
"{sentence}".
\end{promptbox}

\noindent
This simple format provides sufficient context for the pretrained model, without requiring extensive prompt engineering.

\vspace{.5\baselineskip}
\noindent
\textbf{Topical framing} is done using the nine categories described in the schema presented in \autoref{tab:frame-definitions}.
We perform this multiclass classification using \verb|phi-4|~\cite{phi4}.
The model receives an article's title, subtitle (if present), and the full text content. As output, it then provides applicable frames along with confidence scores.

\begin{promptbox}
You are a careful policy text classifier. Your task is to identify the topical frames expressed in the input text using a fixed set of predefined labels. You may assign one or more labels if multiple topics are present.

Use only the following labels: ECONOMY, IDENTITY, LEGAL, MORALITY, POLICY, POLITICS, PUBLIC OPINION, SECURITY, WELFARE.

@\mbox{\textbf{Definitions:}}@
@\textbullet@ ECONOMY: Macroeconomy and markets,
  including sanctions, trade, inflation,
  growth, employment, taxation, and energy
  or commodity prices.
@\textbullet@ IDENTITY: Group identity and belonging,
  including nationality, ethnicity, religion,
  language, gender, discrimination, or
  identity politics.
@\textbullet@ LEGAL: Courts and legal processes,
  including rulings, lawsuits, prosecutions,
  designations, warrants, and legal rights or
  entitlements.
@\textbullet@ MORALITY: Explicit ethical or value-based
  framing, including moral duty, fairness,
  dignity, and humanitarian arguments.
@\textbullet@ POLICY: Concrete government actions,
  including rules, bills, decrees, programs,
  and policy design details describing what a
  policy does.
@\textbullet@ POLITICS: Elections, party or elite
  competition, political strategy, reshuffles,  and coalition dynamics.
@\textbullet@ PUBLIC OPINION: Expressions of public
  sentiment, such as polls, surveys,
  attitudes, or protests.
@\textbullet@ SECURITY: War and defense, violence or
  crime, policing, border enforcement,
  terrorism, or cyberattacks.
@\textbullet@ WELFARE: Social benefits and human services,  including healthcare, pensions, housing aid,  unemployment support, and refugee or
  internally displaced person assistance.

Return STRICT JSON only, with no additional text:
    {
      "labels": [...],
      "confidences": {
        "<LABEL>": <0...1>
      }
    }

If no labels apply, return an empty list for "labels".

---

Article:
  Title: {title}
  Content: {content}
\end{promptbox}

% \vspace{.5\baselineskip}
\noindent
\textbf{Thematic tagging} (see \S\hspace{1pt}\ref{sec:applications}, use case 1: ``A Thematic Study of Conflict Events'') is done by extracting article-level tags using GPT-3.5 Turbo with one-shot prompting.
The model is instructed to extract a set of specific tags summarizing each article’s content.
The tags were intended to capture salient persons or people, locations, military units, political or narrative themes, and ideological frames surrounding major conflict events.

To ensure a consistent structure that can support post-processing, the prompt was instructed to produce outputs in a strict JSON format.

\begin{promptbox}
@\mbox{\textbf{Instruction:}}@
You are a geopolitical analyst trained to extract concise topical tags from news articles related to international conflict, especially the Russo-Ukrainian war.

Your task is to read the content of a single article and extract @\mbox{\textbf{5 to 12 relevant tags}}@ that capture the article's key locations, actors, events, topics, or narrative themes.

Each tag should follow these rules:
- Tags must be @\mbox{\textbf{1-4 words long}}@.
- Use @\mbox{\textbf{lowercase}}@, unless the tag contains a
  @\mbox{\textbf{proper noun}}@.
- Tags should be @\mbox{\textbf{specific and informative}}@,
  not generic.
- Avoid vague tags such as "news", "fake", or
  "war" unless they are central to the
  article.
- Tags may include @\mbox{\textbf{people}}@, @\mbox{\textbf{institutions}}@,
  @\mbox{\textbf{military units}}@, @\mbox{\textbf{locations}}@, @\mbox{\textbf{investigations}}@,
  @\mbox{\textbf{political events}}@, @\mbox{\textbf{legal processes}}@, or
  @\mbox{\textbf{disinformation narratives}}@.

Use @\mbox{\textbf{only the article content}}@ to determine the
tags.

Return only @\mbox{\textbf{valid JSON}}@ with no explanations or additional text.

@\mbox{\textbf{Tagging Guidelines:}}@

Tags may represent:
- Key @\mbox{\textbf{locations}}@ (e.g., Bucha, Donbas, Odessa)
- Key @\mbox{\textbf{actors}}@ (e.g., Sergei Lavrov, Budanov)
- @\mbox{\textbf{Events or incidents}}@ (e.g., Bucha massacre,
  Kramatorsk attack)
- @\mbox{\textbf{Institutions or investigations}}@ (e.g.,
  International Criminal Court, Bellingcat,
  NATO, Wagner)
- @\mbox{\textbf{Narratives or themes}}@ (e.g., false flag
  allegations, media manipulation)
- @\mbox{\textbf{Evidence or technologies}}@ (e.g., satellite
  imagery, intercepted communications)

@\mbox{\textbf{Output Format:}}@
Return the result in valid JSON using the following structure:

    {
      "article_ID": [
        "tag 1",
        "tag 2",
        "tag 3",
        "tag 4",
        "tag 5"
      ]
    }

- Replace "article_ID" with the actual
  article identifier.
- Return @\mbox{\textbf{5-12 tags}}@ per article.
- Do not include explanations or additional
  text.

@\mbox{\textbf{Example 1:}}@

@\mbox{\textbf{Title:}}@ One tragic Ukraine story told with drones, satellites and social media

@\mbox{\textbf{Content:}}@
Tragic individual stories of 21st century war
in Ukraine are coming into focus in 21st
century ways. [...] When Russia invaded,
Filkina stayed behind helping people in Bucha
and cooked for the Ukrainian military. [...]
Der Spiegel reported that Germany's foreign
intelligence agency intercepted Russian radio
chatter about the killing of civilians in
Bucha. [...] Russia is also apparently trying
to tell its own story by hacking into the
social media accounts of Ukrainian soldiers,
according to Meta. [...] The United Nations
voted Thursday to suspend Russia from the UN
Human Rights Council. [...] Organizations
like Bellingcat are using satellite and
social media to document war crimes and
identify perpetrators. [...] International
criminal investigations rely on
authenticated images, witness testimonies,
and other evidence concerning killings,
torture, and crimes against humanity. [...]

@\mbox{\textbf{Example Output:}}@

    {
      "article_example_1": [
        "Bucha killings",
        "Russian invasion",
        "war crimes evidence",
        "social media manipulation",
        "Ukrainian journalists",
        "UN human rights council suspension",
        "war crimes trials",
        "satellite imagery",
        "Bellingcat investigations",
        "International Criminal Court",
        "documentation of atrocities"
      ]
    }

---

Now extract tags from the following article.

Article ID: {{article_id}}
Title: {{title}}
Subtitle: {{subtitle}}
Content: {{content}}

Return only valid JSON.
\end{promptbox}

% \vspace{.5\baselineskip}
\noindent
\textbf{Claim typology} was established with the six types of claims (see \S\hspace{1pt}\ref{sec:applications}, use case 4: ``Divergent Narratives''): \verb|factual|, \verb|attributed|, \verb|disputed|, \verb|reported|, \verb|command/policy|, and \verb|analytical/assessment|.
We use a structured two-shot prompting strategy with GPT-3.5 Turbo to extract and classify claims.

\begin{promptbox}
@\mbox{\textbf{Definition:}}@ A @\mbox{\textbf{claim}}@ is a statement asserting a fact, report, allegation, policy decision, interpretation, or prediction related to the conflict.

@\mbox{\textbf{Tag Definitions:}}@
@\mbox{\textbf{1. Factual:}}@ A statement presented as an
   objective fact without attribution to a
   specific actor or source.
@\mbox{\textbf{2. Attributed:}}@ A claim explicitly attributed
   to a person, organization, government, or
   authority.
@\mbox{\textbf{3. Disputed:}}@ A claim that is challenged,
   denied, or contradicted by another actor
   or source.
@\mbox{\textbf{4. Reported:}}@ A claim framed as being reported
   by media, unnamed officials, or sources
   without direct confirmation.
@\mbox{\textbf{5. Command/Policy:}}@ A directive, official
   order, government decision, or announced
   policy action.
@\mbox{\textbf{6. Analytical/Assessment:}}@ Interpretation,
   analysis, speculation, prediction, or
   expert evaluation about the conflict.

@\mbox{\textbf{Instruction:}}@
You are a geopolitical analyst trained to extract claims from news articles about the Russia-Ukraine war.

Your task is to read the content of a single article and extract every distinct claim explicitly stated in the text.

Follow these rules carefully:
- Extract only claims explicitly present in
  the article text.
- Do not invent or infer claims not supported
  by the text.
- Claims may span one or multiple consecutive
  sentences if they form a single coherent
  statement.
- Do not split a single claim unnecessarily.
- Do not merge unrelated statements into one
  claim.
- Preserve the exact wording of the original
  sentence(s) in the "sentence" field.
- The "summary" must be concise, neutral, and
  factual.
- Use ONLY the article content for analysis.
- Extract all distinct claims present in the
  article.
- Return only valid JSON and no additional
  explanation.

@\mbox{\textbf{Output Format:}}@
Return the result in the following JSON structure:

    {
      "article_ID": {
        "claim_1": {
          "sentence": "Original sentence(s) from the article expressing the claim.",
          "summary": "Short neutral summary of the claim.",
          "claim_type": "One of the predefined claim types"
        }
      }
    }

- Replace "article_ID" with the actual
  article identifier.
- Number claims sequentially (claim_1,
  claim_2, ...).
- The "sentence" field must contain the
  original sentence(s) from the article.
- The summary must be concise and neutral.
- The claim_type must match exactly one of
  the predefined categories.
- Return only JSON with no additional text
  or explanations.

@\mbox{\textbf{Example 1:}}@

@\mbox{\textbf{Title:}}@ Russia Hits Macron, Shifting Political
       Dynamics Ahead of Elections in France  

@\mbox{\textbf{Content:}}@
48.7 million French voters are officially registered to cast their ballot [...] Emmanuel Macron announced that a new round of sanctions against Russia was needed and said there are very clear clues pointing to war crimes in Bucha. [...] Green party's Yannick Jadot called for a full embargo on Russian gas and oil, criticising Macron's relationship with Russian President Vladimir Putin. [...] Support for Macron has now fallen to 27%, according to Politico's Poll of Polls, while Marine Le Pen's voting intentions have surged from 17% to 22% since the start of the war. [...]

@\mbox{\textbf{Example Output:}}@
{
  "article_example_1": {
    "claim_1": {
      "sentence": "Green party's Yannick
      Jadot called on Thursday for a full
      embargo on Russian gas and oil,
      criticising Macron's seemingly-warm
      relationship with Russian President
      Vladimir Putin.",
      "summary": "Jadot calls for embargo on
      Russian gas and oil and criticizes
      Macron's stance on Putin.",
      "claim_type": "Attributed"
    },
    "claim_2": {
      "sentence": "Support has now fallen to
      27%, according to Politico's Poll of
      Polls. Marine Le Pen has seen voting
      intentions in her favour surge from 17%
      to 22% since the start of the war.",
      "summary": "Macron's support decreases
      to 27% while Le Pen's rises to 22% in
      the polls.",
      "claim_type": "Reported"
    },
    "claim_3": {
      "sentence": "Emmanuel Macron announced
      that a new round of sanctions against
      Russia was needed and said there are
      very clear clues pointing to war crimes
      in Bucha.",
      "summary": "Macron calls for new
      sanctions against Russia and accuses
      them of war crimes in Bucha.",
      "claim_type": "Command/Policy"
    }
  }
}

@\mbox{\textbf{Example 2:}}@

@\mbox{\textbf{Title:}}@ The horror of the Kramatorsk station
       attack

@\mbox{\textbf{Content:}}@
The Russian missile attack on Kramatorsk railway station that killed at least 50 people [...] It will fuel suspicions that Russia is targeting the civilian population to sow fear and break resistance after the setbacks its army has experienced in six weeks of war. [...] The reaction of Moscow officials and media followed a pattern of obfuscation, denial and outright fiction. [...] The different versions echoed the conflicting stories Russia used to try to distance itself from the downing in 2014 of Malaysia Airlines flight MH17. [...]

@\mbox{\textbf{Example Output:}}@
{
  "article_example_2": {
    "claim_1": {
      "sentence": "It will fuel suspicions
      that Russia is targeting the civilian
      population to sow fear and break resis-
      tance, after the setbacks its army has
      experienced in six weeks of war.",
      "summary": "The attack may reflect a
      strategy of targeting civilians to
      intimidate Ukraine.",
      "claim_type": "Analytical/Assessment"
    },
    "claim_2": {
      "sentence": "The different versions
      echoed the conflicting stories Russia
      used to try to distance itself from the
      downing in 2014 of Malaysia Airlines
      flight MH17.",
      "summary": "Russia has previously used
      conflicting narratives to distance itself
      from incidents like the MH17 downing.",
      "claim_type": "Reported"
    }
  }
}

---

Now extract claims from the following article.

Article ID: {{article_id}}
Title: {{title}}
Subtitle: {{subtitle}}
Content: {{content}}

Return only valid JSON.
\end{promptbox}

\section*{Appendix B. Use Case Experiments}
\label{app-b:use-case-experiments}
% \vspace*{.5\baselineskip}
% \noindent
% \textbf{Temporal Sentiment Analysis}

\begin{figure*}
    \centering
    \includegraphics[width=\textwidth]{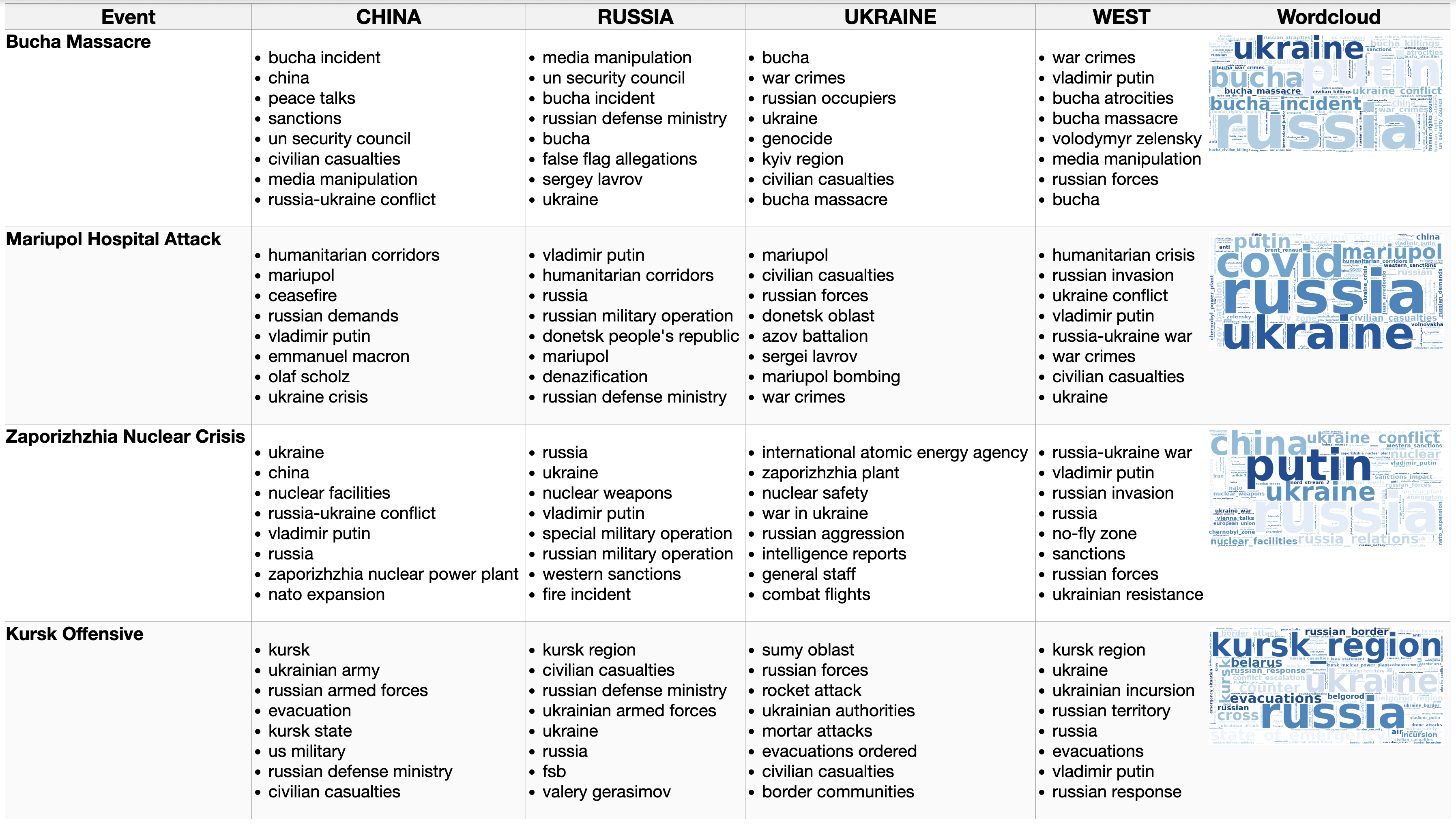}
    \caption{Thematic emphasis patterns across news outlets for four major conflict events, derived from GPT-3.5-tagged concepts, actors, and subtopics aggregated by outlet country.}
    \label{fig:word-clouds}
\end{figure*}

\paragraph{Thematic Study of Conflict Events:}
We present word clouds in \autoref{fig:word-clouds} to visually support our thematic analysis (described in \S\hspace{1pt}~\ref{sec:applications}. 
Using GPT-3.5~\cite{openai2023gpt35} with structured prompts, we tag articles from weekly time windows surrounding each major event to identify key concepts, actors, and subtopics. For the events described in \autoref{tab:use-case-events}, we obtain 810 articles, with 10.5 such tags (avg.) per article. We aggregate these tags by event and country to find the distinct thematic emphasis patterns across media outlets.

In summary, each word cloud aggregates GPT-3.5-tagged concepts, actors, and subtopics by outlet country for each of the four conflict events, with term size reflecting relative frequency. These visualizations highlight the varied thematic focus across geopolitical actors.

\begin{figure*}[!t]
\centering
\includegraphics[width=0.48\linewidth]{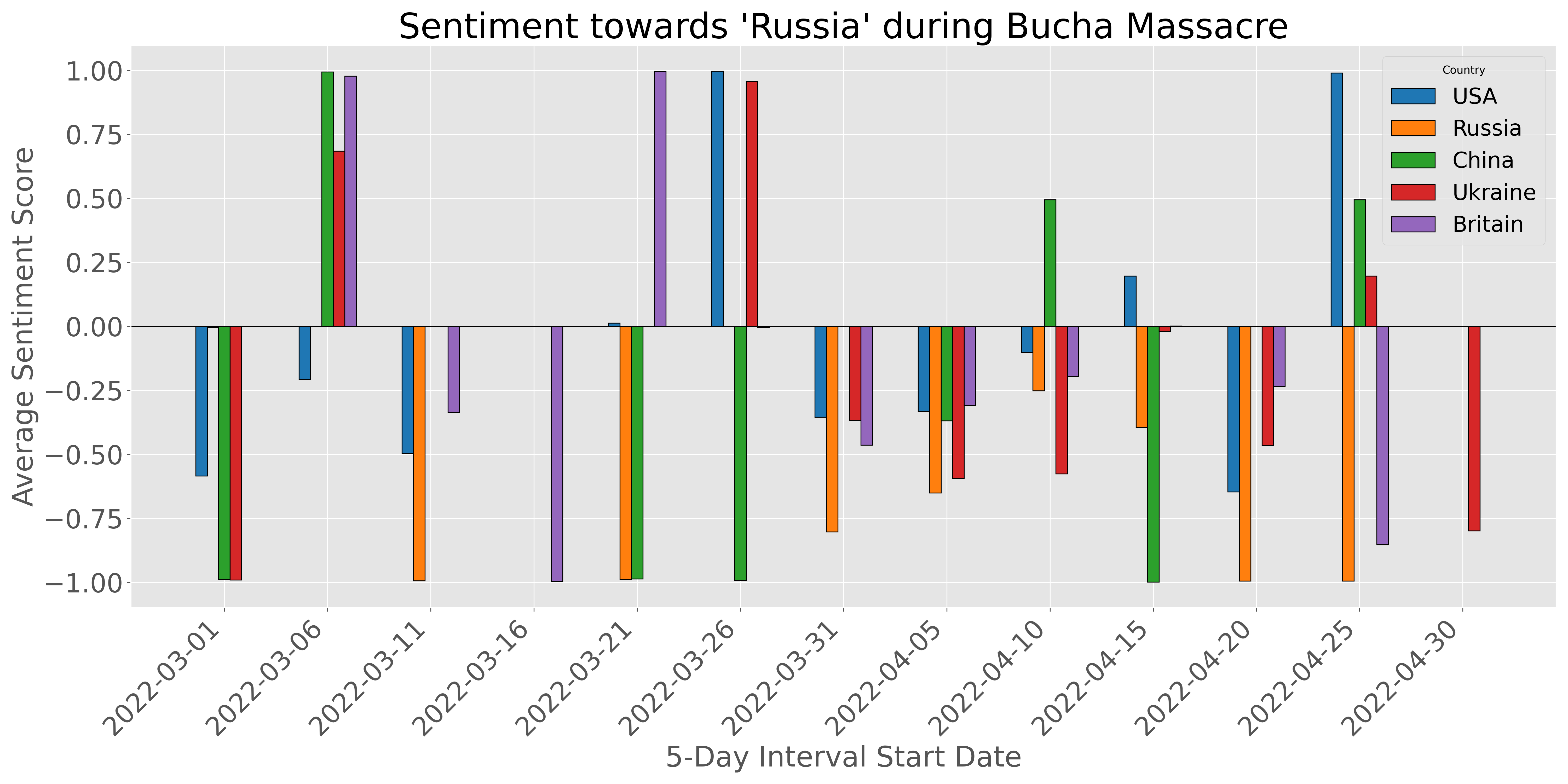}
\hfill%
\includegraphics[width=0.48\linewidth]{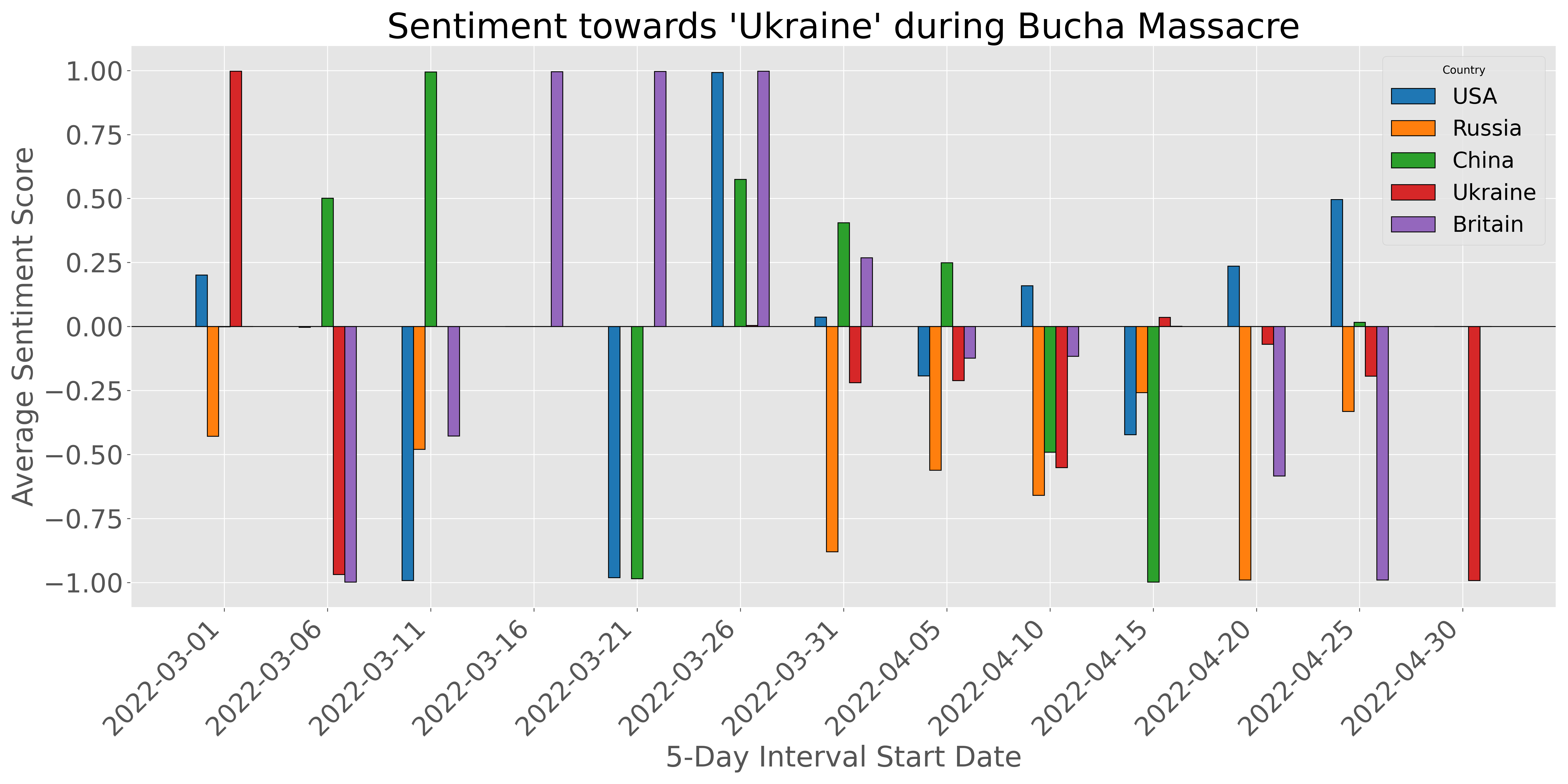}
\caption{\small The average sentiment scores at 5-day intervals, for Russia (left) and Ukraine (right), during the aftermath of the Bucha massacre (March - April 2022). News outlets are grouped by their national affiliations.}
\label{fig:sentiment_russia_ukraine}
\end{figure*}

\paragraph{Temporal Sentiment Analysis:}
\autoref{fig:sentiment_russia_ukraine} shows 5-day rolling average sentiment scores toward Russia and Ukraine in the immediate aftermath of the Bucha massacre (Mar - Apr 2022), aggregated by source country.
Ukrainian media exhibits consistently negative sentiment toward Russia, as expected given their direct involvement in the conflict.
Occasional positive spikes likely reflect optimistic coverage of military resistance or international support pledges.
Russian media also shows a deeply negative sentiment -- counterintuitive, given state alignment.
This pattern suggests that war reporting employs vocabulary (violence, casualties, destruction) that sentiment models flag as inherently negative, regardless of political framing.

Media in the USA and the UK exhibit pronounced volatility, with positive sentiment spikes corresponding to perceived diplomatic wins such as the US President's Warsaw speech or high-level visits to Kyiv. Chinese media maintains characteristically moderate sentiment, consistent with their ``strategic partnership'' framing, though scores decline during peak media coverage of Bucha in early April.

These patterns illustrate how sentiment analysis captures both explicit editorial stance and the underlying semantic content of war reporting, which may skew negative even if events are framed favorably.

\section*{Appendix C. Annotation Guidelines}
\label{app-c:human-annotation-guidelines}

We share the guidelines provided to human annotators for the validation experiments (see \S\hspace{1pt}\ref{ssec:human-evaluation}).

\vspace{.5\baselineskip}
\noindent\textbf{Topical Framing Annotation:}
The \textbf{task} is to identify the topical frames expressed in an article using the full text as context.
Articles may receive multiple labels reflecting co-occurring themes.
The \textbf{labels} are defined as per \autoref{tab:frame-definitions}.
Annotators are instructed as follows:
\begin{itemize}[leftmargin=*, nolistsep, noitemsep]
\item assign labels based on explicit or implicit framing cues in the text;
\item multiple labels are permitted and encouraged when themes co-occur;
\item provide confidence scores \verb|[0, 1]| for each assigned label; and
\item avoid ranking labels by importance.
\end{itemize}

\vspace*{.5\baselineskip}
\noindent
\textbf{Stance Annotation:} The \textbf{task} is to determine the stance expressed in a sentence toward a specified target, using the surrounding text as context.
Each sentence receives exactly one of these \textbf{labels}:
\begin{itemize}[leftmargin=18pt, nolistsep, noitemsep]
\item[(1)] \textsc{favor}: positive stance (support, praise, endorsement);
\item[(2)] \textsc{against}: negative stance (criticism, blame, opposition); or
\item[(3)] \textsc{none}: No clear stance, or purely descriptive text, or ambiguous stance.
\end{itemize}
The annotators are provided these instructions:
\begin{itemize}[leftmargin=*, nolistsep, noitemsep]
\item \textit{Use the full passage} to resolve ambiguous pronouns or implicit references.
\item If the target is \textsc{none}, the stance is also \textsc{none}.
\item Focus on the stance expressed in the sentence, not your personal opinion.
\item Implicit stances (e.g., through selective vocabulary) should be labeled.
\end{itemize}
\autoref{tab:topical_examples} and \autoref{tab:stance_examples} provide annotated examples from the gold-standard human-annotated validation set, illustrating the topical framing and stance annotation tasks, respectively.
For stance annotation, each sentence is paired with a stance label assigned toward a target entity.
For topical framing, each passage may be associated with one or more topical frames, reflecting the co-occurrence of themes within the same article.
The highlighted spans indicate the textual evidence that motivated the assigned labels.
These highlights are intended to make the annotation rationale explicit and to ease reader interpretation; they were not provided to human annotators, and as such, the annotations are \textit{not} derived from any span-level supervision.

{
\begin{table*}[htbp]
\centering
\small
\begin{tabularx}{\textwidth}{@{}X p{.15\textwidth}@{}}
\toprule
\textbf{Passage} & \textbf{Frames}\medskip\\
% \textbf{Title:} Protests at Washington University peaceful, no arrests — police.\newline%
% \textbf{Excerpt:} \textit{According to a spokesman for the city police department, the Metropolitan Police Department, supported by the District of Columbia, and the George Washington University Police, have been monitoring the situation since April 25.} 
% & \textsc{public opinion} (\texttt{0.9})\newline%
% \textsc{security} (\texttt{0.6})\smallskip\\
\textbf{Title:} \textit{Putin gives honorary title to Russian brigade accused of war crimes in Bucha}\vspace{.5\baselineskip}\newline%
\textbf{Excerpt:} \textit{A Russian brigade accused of committing war crimes in Bucha was awarded an honorary title by President Vladimir Putin. Putin congratulated the unit for its "great heroism and courage" and awarded it the title of "Guards" for "protecting Russia's sovereignty." International leaders condemned the alleged atrocities, while investigators cited "reasonable grounds" to believe war crimes were committed.} 
& $\bullet$ \textsc{morality}: \texttt{0.9}\newline%
$\bullet$ \textsc{politics}: \texttt{0.6}\newline%
$\bullet$ \textsc{legal}: \texttt{0.6}\newline%
$\bullet$ \textsc{security}: \texttt{0.6}\\
\bottomrule
\end{tabularx}
\caption{\small Topical framing annotation examples showing multi-label assignments with confidence scores.}
\label{tab:topical_examples}
\end{table*}
}

{
\setlength{\tabcolsep}{12pt}
\begin{table*}[htbp]
\centering
\small
\begin{tabularx}{\linewidth}{@{}X r@{}}
\toprule
\textbf{Sentence (with \yh{target})} & \textbf{Stance}\medskip\\
Russia has failed to capture Kyiv or Kharkiv, Ukraine's two biggest cities, and suffered heavy losses against a far stronger \yh{\textbf{Ukraine}} than even its allies expected.\smallskip& \textsc{favor}\\
Aurel Braun, a professor at the University of Toronto who has written presciently on how western governments and companies misjudged the irrationalities of the \yh{\textbf{Putin regime}}, points out that international relations is full of theories about how the interplay of powerful corporate interests and government produces policy. \smallskip& \textsc{against}\\
When we see \yh{\textbf{Russia}}'s unprovoked and unjustified invasion of Ukraine and massive disinformation campaigns and information manipulation, it is essential to separate lies from facts.\smallskip& \textsc{against}\\
US officials have calculated that \yh{\textbf{Putin}}, facing crippling sanctions across other sectors of his economy, can't afford to weaponize his energy supplies by cutting off sales to Europe.\smallskip& \textsc{none}\\
\yh{\textbf{Vladimir Putin}} awarded Heroes of the Russian Federation in the Yekaterininsky Hall.& \textsc{none}\\
\bottomrule
\end{tabularx}
\caption{\small Stance annotations from the gold-standard human validation set, shown with highlighted target entities.}
\label{tab:stance_examples}
\end{table*}
}

\end{document}